\documentclass[review]{elsarticle}
\biboptions{longnamesfirst,semicolon}

\usepackage{url}
\usepackage{latexsym}
\usepackage{color}
\usepackage{multirow}
\usepackage{booktabs}
\usepackage{longtable}
\usepackage{enumitem}
\usepackage{graphicx}
\usepackage{rotating}

\definecolor{orange}{rgb}{1,0.5,0}

\begin{document}

\title{Generalisation in Named Entity Recognition: A Quantitative Analysis}

\author{Isabelle Augenstein}
\ead{i.augenstein@ucl.ac.uk}
\author{Leon Derczynski}
\author{Kalina Bontcheva}

\address{University of Sheffield, Sheffield, S1 4DP, UK}

\begin{abstract}
Named Entity Recognition (NER) is a key NLP task, which is all the more challenging on Web and user-generated content with their diverse and continuously changing language.
This paper aims to quantify how this diversity impacts state-of-the-art NER methods, by measuring named entity (NE) and context variability, feature sparsity, and their effects on precision and recall. In particular, our findings indicate that NER approaches struggle to generalise in diverse genres with limited training data.
Unseen NEs, in particular, play an important role, which have a higher incidence in diverse genres such as social media than in more regular genres such as newswire. Coupled with a higher incidence of unseen features more generally and the lack of large training corpora, this leads to significantly lower F1 scores for diverse genres as compared to more regular ones.
We also find that leading systems rely heavily on surface forms found in training data, having problems generalising beyond these, and offer explanations for this observation.

\end{abstract}

\maketitle

\section{Introduction}

Named entity recognition and classification (\textit{NERC}, short \textit{NER}), the task of recognising and assigning a class to mentions of proper names (named entities, \textit{NEs}) in text, has attracted many years of research \cite{nadeau2007survey,RatinovRo09}, analyses \cite{Palmer1997Statistical}, starting from the first MUC challenge in 1995~\citep{grishman_sundheim_1995}. Recognising entities is key to many applications, including text summarisation~\cite{schiffman2002experiments}, search~\cite{toda2005search}, the semantic web~\cite{maynard2016natural}, topic modelling~\cite{newman2006statistical}, and machine translation~\cite{al2002translating,steinberger2011jrc}.

%Despite high precision and recall being reported on static benchmark datasets (predominantly newswire), NER systems struggle with content where the named entities mentioned change over time (\emph{entity drift}), such as Web and user-generated content~\citep{Derczynski2013b,fromreide2014crowdsourcing}.

%It is therefore important to recognise where as a field research on NER is headed, and what ground has been covered. 
%We are facing a large number of, and more diverse, corpora and genres, and the field has undergone a shift from newswire-related entity extraction, to having to recognise and classify a much more diverse range of entities for a broad range of applications. 
%This leads to a noisier, sparser feature space, including many more linguistic phenomena and styles. 
%However, while there is a body of excellent work on domain adaptation~\cite{DaumeIII2007Domain,wu2009domain,guo-EtAl:2009:NAACLHLT09,chiticariu2010domain} and specialist NER~\cite{tanabe2005genetag,li2012twiner}, general assumptions in the field have not necessarily been re-thought to account for the significantly increased scope of modern named entity recognition.

As NER is being applied to increasingly diverse and challenging text genres~\cite{whitelaw2008web,Derczynski2014b,fromreide2014crowdsourcing}, this has lead to a noisier, sparser feature space, which in turn requires regularisation~\cite{cherryunreasonable} and the avoidance of overfitting. This has been the case even for large corpora all of the same genre and with the same entity classification scheme, such as ACE~\cite{mooney2005subsequence}.
Recall, in particular, has been a persistent problem, as named entities often seem to have unusual surface forms, e.g. unusual character sequences for the given language (e.g. {\em Szeged} in an English-language document) or words that individually are typically not NEs, unless they are combined together (e.g. {\em the White House}). 
%Gazetteers often play a key role in overcoming low NER recall~\cite{Mikheev99}.
%These are sets of phrases that have a property in common, for example, a gazetteer of capital city names in English, or a gazetteer of military titles, would be a list of each of those things.
%Research on gazetteer collection has moved from manual assembly~\cite{cunningham-EtAl:2002:ACL}, through automatic collection~\cite{kozareva2006bootstrapping,maynard2009sprat}, and now the most recent NER challenges have distributed entity gazetteers~\cite{baldwin2015shared} derived from Linked Data~\cite{bollacker2008freebase} as part of their baseline systems.

Indeed, the move from ACE and MUC to broader kinds of corpora has presented existing NER systems and resources with a great deal of difficulty~\cite{MBC03}, which some researchers have tried to address through domain adaptation, specifically with entity recognition in mind~\cite{DaumeIII2007Domain,wu2009domain,guo-EtAl:2009:NAACLHLT09,chiticariu2010domain,augenstein2014joint}.
However, more recent performance comparisons of NER methods over different corpora showed that older tools tend to simply fail to adapt, even when given a fair amount of in-domain data and resources~\cite{ritter2011named,Derczynski2014b}.
Simultaneously, the value of NER in non-newswire data~\citep{ritter2011named,liu2011recognizing,plank2014adapting,rowe2015microposts2015,baldwin2015shared} has rocketed: for example, social media now provides us with a sample of all human discourse, unmolested by editors, publishing guidelines and the like, and all in digital format -- leading to, for example, whole new fields of research opening in computational social science~\cite{hovy2015user,plank2015personality,preoctiuc2015studying}.

%Two major perceived challenges are that NEs mentioned in tweets change over time (the drift)~\citep{Derczynski2013b,fromreide2014crowdsourcing}, and that diversity of context makes NER more difficult~\citep{Derczynski2014b}.

The prevailing assumption has been that this lower NER performance is due to domain differences arising from using newswire (NW) as training data, as well as from the irregular, noisy nature of new media (e.g.~\cite{ritter2011named}). Existing studies~\citep{Derczynski2014b} further suggest that named entity diversity, discrepancy between named entities in the training set and the test set (\emph{entity drift} over time in particular), and diverse context, are the likely reasons behind the significantly lower NER performance on social media corpora, as compared to newswire. 

%This paper, however, argues that overfitting has occurred not only at dataset and model level, but also through the community's reliance on ageing, low-diversity, predominantly newswire datasets.

No prior studies, however, have investigated these hypotheses quantitatively. For example, it is not yet established whether this performance drop is really due to a higher proportion of unseen NEs in the social media, or is it instead due to NEs being situated in different kinds of linguistic context.

Accordingly, the contributions of this paper lie in investigating the following open research questions:

\begin{description}[noitemsep]
%How does NERC performance differ for corpora over different genres?

\item[RQ1] How does NERC performance differ for corpora between different NER approaches?
\item[RQ2] How does NERC performance differ for corpora over different text types/genres?
\item[RQ3] What is the impact of NE diversity on system performance?
\item[RQ4] What is the relationship between Out-of-Vocabulary (OOV) features (unseen features), OOV entities (unseen NEs) and performance?
\item[RQ5] How well do NERC methods perform out-of-domain and what impact do unseen NEs (i.e. those which appear in the test set, but not the training set) have on out-of-domain performance?
\end{description}

In particular, the paper carries out a comparative analyses of the performance of several different approaches to statistical NER over multiple text genres, with varying NE and lexical diversity.
In line with prior analyses of NER performance~\cite{Palmer1997Statistical,Derczynski2014b}, we carry out corpus analysis and introduce briefly the NER methods used for experimentation.
Unlike prior efforts, however, our main objectives are to uncover the impact of NE diversity and context diversity on performance (measured primarily by F1 score), and also to study the relationship between OOV NEs and features and F1. See Section~\ref{sec:experiments} for details.

To ensure representativeness and comprehensiveness, our experimental findings are based on key benchmark NER corpora spanning multiple genres, time periods, and corpus annotation methodologies and guidelines. As detailed in Section~\ref{sec:datasets}, the corpora studied are OntoNotes~\citep{Hovy2006OntoNotes}, ACE~\citep{walker2006ace}, MUC 7~\citep{chinchor98MUC}, the Ritter NER corpus~\citep{ritter2011named}, the MSM 2013 corpus~\citep{rowe2013msm}, and the UMBC Twitter corpus~\cite{finin-EtAl:2010:MTURK}. To eliminate potential bias from the choice of statistical NER approach, experiments are carried out with three differently-principled NER approaches, namely Stanford NER~\cite{Finkel2005StanfordNER}, SENNA~\cite{Collobert2011Senna} and CRFSuite~\cite{okazaki2007crfsuite} (see Section~\ref{sec:models} for details).

%We address this these questions by first describing the state of the art as context (Section~\ref{sec:related_work}), and going on to use a diverse range of datasets and technologies (Section~\ref{sec:data_methods}) in Section~\ref{sec:experiments}'s experiments. The results and analyses are summed up in our conclusion, Section~\ref{sec:conclusion}.

%\section{Related Work} \label{sec:related_work}

\setlength{\tabcolsep}{0.3em}
\begin{table*}[t]
\fontsize{9}{11}\selectfont
\centering
\begin{tabular}{c|c|c|c c c}
\toprule
\textbf{Corpus} & \textbf{Genre} & \textbf{N} & \textbf{PER} & \textbf{LOC} & \textbf{ORG} \\
\hline
MUC 7 Train & Newswire (NW) & 552 & 98 & 172 & 282 \\
MUC 7 Dev & Newswire (NW) & 572 & 93 & 193 & 286 \\
MUC 7 Test & Newswire (NW) & 863 & 145 & 244 & 474 \\
\hline
CoNLL Train & Newswire (NW) & 20061 & 6600 & 7140 & 6321 \\
CoNLL TestA & Newswire (NW) & 4229 & 1641 & 1434 & 1154 \\
CoNLL TestB & Newswire (NW) & 4946 & 1617 & 1668 & 1661 \\
\hline
ACE NW & Newswire (NW) & 3835 & 894 & 2238 & 703 \\
ACE BN & Broadcast News (BN) & 2067 & 830 & 885 & 352 \\
ACE BC & Broadcast Conversation (BC) & 1746 & 662 & 795 & 289 \\
ACE WL & Weblog (WEB) & 1716 & 756 & 411 & 549 \\
ACE CTS & Conversational Telephone Speech (CTS) & 2667 & 2256 & 347 & 64 \\
ACE UN & Usenet Newsgroups (UN) & 668 & 277 & 243 & 148 \\
\hline
OntoNotes NW & Newswire (NW) & 52055 & 12640 & 16966 & 22449 \\
OntoNotes BN & Broadcast News (BN) & 14213 & 5259 & 5919 & 3035 \\
OntoNotes BC & Broadcast Conversation (BC) & 7676 & 3224 & 2940 & 1512 \\
OntoNotes WB & Weblog (WEB) & 6080 & 2591 & 2319 & 1170 \\
OntoNotes TC & Telephone Conversations (TC) & 1430 & 745 & 569 & 116 \\
OntoNotes MZ & Magazine (MZ) & 8150 & 2895 & 3569 & 1686 \\
\hline
MSM 2013 Train & Twitter (TWI) & 2815 & 1660 & 575 & 580 \\
MSM 2013 Test & Twitter (TWI) & 1432 & 1110 & 98 & 224 \\
Ritter & Twitter (TWI) & 1221 & 454 & 380 & 387 \\
UMBC & Twitter (TWI) & 510 & 172 & 168 & 170 \\
\bottomrule
\end{tabular}
\caption{\label{tab:CorpTypes} Corpora genres and number of NEs of different classes}
\end{table*}

\section{Datasets and Methods}\label{sec:data_methods}

\subsection{Datasets}\label{sec:datasets}

Since the goal of this study is to compare NER performance on corpora from diverse domains and genres, seven benchmark NER corpora are included, spanning newswire, broadcast conversation, Web content, and social media (see Table~\ref{tab:CorpTypes} for details). 
These datasets were chosen such that they have been annotated with the same or very similar entity classes, in particular, names of people, locations, and organisations. 
Thus corpora including only domain-specific entities (e.g. biomedical corpora) were excluded. The choice of corpora was also motivated by their chronological age; we wanted to ensure a good temporal spread, in order to study possible effects of entity drift over time.\footnote{Entity drift is a cause of heightened diversity. It happens when the terms used to represent an NE category, such as ``person", change over time. For example, ``yeltsin" might be one lexicalisation of a person entity in the 1990s, whereas one may see ``putin" in later texts. Both are lexical representations of the same kind of underlying entity, which has remained static over time. Entity drift is a form of concept drift~\cite{masud2010addressing} specific to NER.}

A note is required about terminology.
This paper refers to text {\em genre} and also text {\em domain}.
These are two dimensions by which a document or corpus can be described.
Genre here accounts the general characteristics of the text, measurable with things like register, tone, reading ease, sentence length, vocabulary and so on.
Domain describes the dominant subject matter of text, which might give specialised vocabulary or specific, unusal word senses.
For example, ``broadcast news" is a genre, describing the manner of use of language, whereas ``financial text" or ``popular culture" are domains, describing the topic.
One notable exception to this terminology is social media, which tends to be a blend of myriad domains and genres, with huge variation in both these dimensions~\cite{Hu2013,baldwin2013noisy}; for simplicity, we also refer to this as a genre here.

\subsubsection{Corpora Used} \label{sec:corpora}

In chronological order, the first corpus included here is MUC 7, which is the last of the MUC challenges~\cite{chinchor98MUC}. This is an important corpus, since the Message Understanding Conference (MUC) was the first one to introduce the NER task in 1995~\citep{grishman_sundheim_1995}, with focus on recognising persons, locations and organisations in newswire text.

A subsequent evaluation campaign was the CoNLL 2003 NER shared task~\citep{tjong2003introduction}, which created gold standard data for newswire in Spanish, Dutch, English and German. The corpus of this evaluation effort is now one of the most popular gold standards for NER, with new NER approaches and methods often reporting performance on that.

Later evaluation campaigns began addressing NER for genres other than newswire, specifically ACE~\citep{walker2006ace} and OntoNotes~\citep{Hovy2006OntoNotes}. Both of those contain subcorpora in several genres, namely newswire, broadcast news, broadcast conversation, weblogs, and conversational telephone speech. ACE, in addition, contains a subcorpus with usenet newsgroups. Like CoNLL 2003, the OntoNotes corpus is also a popular benchmark dataset for NER. The languages covered are English, Arabic and Chinese. A further difference between the ACE and OntoNotes corpora on one hand, and CoNLL and MUC on the other, is that they contain annotations not only for NER, but also for other tasks such as coreference resolution, relation and event extraction and word sense disambiguation. In this paper, however, we restrict ourselves purely to the English NER annotations, for consistency across datasets. The ACE corpus contains HEAD as well as EXTENT annotations for NE spans. For our experiments we use the EXTENT tags.

With the emergence of social media, studying NER performance on this genre gained momentum. So far, there have been no big evaluation efforts, such as ACE and OntoNotes, resulting in substantial amounts of gold standard data. Instead, benchmark corpora were created as part of smaller challenges or individual projects. The first such corpus is the UMBC corpus for Twitter NER \cite{finin-EtAl:2010:MTURK}, where researchers used crowdsourcing to obtain annotations for persons, locations and organisations. A further Twitter NER corpus was created by~\cite{ritter2011named}, which, in contrast to other corpora, contains more fine-grained classes defined by the Freebase schema~\citep{bollacker2008freebase}.
 Next, the Making Sense of Microposts initiative~\citep{rowe2013msm} (MSM) provides single annotated data for named entity recognition on Twitter for persons, locations, organisations and miscellaneous. MSM initiatives from 2014 onwards in addition feature a named entity linking task, but since we only focus on NER here, we use the 2013 corpus.

These corpora are diverse not only in terms of genres and time periods covered, but also in terms of NE classes and their definitions. In particular, the ACE and OntoNotes corpora try to model entity metonymy by introducing facilities and geo-political entities (GPEs). Since the rest of the benchmark datasets do not make this distinction, metonymous entities are mapped to a more common entity class (see below).  

In order to ensure consistency across corpora, only Person (PER), Location (LOC) and Organisation (ORG) are used in our experiments,  and other NE classes are mapped to O (no NE). For the Ritter corpus, the 10 entity classes are collapsed to three as in~\cite{ritter2011named}.
For the ACE and OntoNotes corpora, the following mapping is used: PERSON $\rightarrow$ PER; LOCATION, FACILITY, GPE $\rightarrow$ LOC; ORGANIZATION $\rightarrow$ ORG; all other classes $\rightarrow$ O.

Tokens are annotated with BIO sequence tags, indicating that they are the beginning (B) or inside (I) of NE mentions, or outside of NE mentions (O). For the Ritter and ACE 2005 corpora, separate training and test corpora are not publicly available, so we randomly sample 1/3 for testing and use the rest for training. The resulting training and testing data sizes measured in number of NEs are listed in Table~\ref{tab:CorpSizesAll}. Separate models are then trained on the training parts of each corpus and evaluated on the development (if available) and test parts of the same corpus. If development parts are available, as they are for CoNLL (CoNLL Test A) and MUC (MUC 7 Dev), they are not merged with the training corpora for testing, as it was permitted to do in the context of those evaluation challenges.

\setlength{\tabcolsep}{0.3em}
\begin{table*}[t]
\fontsize{9}{11}\selectfont
\centering
\begin{tabular}{c|c c c}
\toprule
\textbf{Corpus} & \textbf{Train} & \textbf{Dev} & \textbf{Test} \\
\hline
MUC 7 & 552 & 572 & 863 \\
ConLL Train & 20061 & 4229 & 4964 \\
ACE NW & 2624 &  & 1211 \\
ACE BN & 1410 &  & 657 \\
ACE BC & 1157 &  & 589 \\
ACE WL & 954 &  & 762 \\
ACE CTS & 1639 &  & 1028 \\
ACE UN & 354 &  & 314 \\
OntoNotes NW & 33676 &  & 18379 \\
OntoNotes BN & 8316 &  & 5897 \\
OntoNotes BC & 5526 &  & 2150 \\
OntoNotes WB & 3850 &  & 2230 \\
OntoNotes TC & 856 &  & 574 \\
OntoNotes MZ & 5853 &  & 2297 \\
MSM 2013 & 2815 &  & 1432 \\
Ritter 2013 & 816 &  & 405 \\
UMBC & 321 &  & 189 \\
\bottomrule
\end{tabular}
\caption{\label{tab:CorpSizesAll} Sizes of corpora, measured in number of NEs, used for training and testing. Note that the for the ConLL corpus the dev set is called ``Test A'' and the test set ``Test B''.}
\end{table*}

\setlength{\tabcolsep}{0.2em}
\begin{sidewaystable*}[t]
\fontsize{9}{11}\selectfont
\centering
\begin{tabular}{c c|c c c|c c c|c c c|c c c}
\toprule
&  & \multicolumn{3}{c}{\bf CRFSuite} & \multicolumn{3}{c}{\bf Stanford NER} & \multicolumn{3}{c}{\bf SENNA} & \multicolumn{3}{c}{\bf Average}\\
\bf Corpus & \bf Genre & \bf P & \bf R & \bf F1 & \bf P & \bf R & \bf F1 & \bf P & \bf R & \bf F1  & \bf P & \bf R & \bf F1 \\
\hline
MUC 7 Dev & NW & 62.95 & 61.3 & 62.11 & 68.94 & 69.06 & 69 & 55.82 & 65.38 & 60.23 & 62.6 & 65.2 & 63.8\\
MUC 7 Test & NW & 63.4 & 50.17 & 56.02 & 70.91 & 51.68 & 59.79 & 54.35 & 51.45 & 52.86 & 62.9 & 51.1 & 56.2 \\
\hline
CoNLL TestA & NW & 66.63 & 44.62 & 53.45 & 70.31 & 48.1 & 57.12 & 72.22 & 70.75 & 71.48 & 69.7 & 54.5 & 60.7 \\
CoNLL TestB & NW & 67.73 & 43.47 & 52.95 & 69.61 & 44.88 & 54.58 & 48.6 & 48.46 & 48.53 & 62.0 & 45.6 & 52.0 \\
\hline
ACE NW & NW & \textbf{49.73} & 30.72 & 37.98 & \textbf{46.41} & 34.19 & 39.37 & 46.78 & 50.45 & 48.55 & 47.6 & 38.5 & 42.0 \\
ACE BN & BN & 56.69 & \textbf{13.55} & \textbf{21.87} & \textbf{56.09} & \textbf{26.64} & \textbf{36.12} & 40.07 & \textbf{36.83} & \textbf{38.38} & 51.0 & 25.7 & 32.1 \\
ACE BC & BC & 59.46 & 29.88 & 39.77 & 60.51 & 40.07 & 48.21 & \textbf{39.46} & \textbf{41.94} & \textbf{40.66} & 53.1 & 37.3 & 42.9 \\
ACE WL & WEB & 65.48 & \textbf{21.65} & 32.54 & 59.52 & \textbf{22.57} & \textbf{32.73} & 53.07 & \textbf{32.94} & \textbf{40.65} & 59.4 & 25.7 & 35.3 \\
ACE CTS & TC & 69.77 & \textbf{14.61} & \textbf{24.15} & 74.76 & \textbf{23.05} & \textbf{35.24} & 72.36 & 68 & 70.11 & 72.3 & 35.2 & 43.2 \\
ACE UN & UN & \textbf{20} & \textbf{0.41} & \textbf{0.81} & \textbf{10.81} & \textbf{1.65} & \textbf{2.87} & \textbf{12.59} & \textbf{7.44} & \textbf{9.35} & 14.5 & 3.2 & 4.3 \\
\hline
OntoNotes NW & NW & 53.48 & 28.42 & 37.11 & 64.03 & 30.45 & 41.28 & \textbf{36.84} & 51.56 & 42.97 & 51.5 & 36.8 & 40.5 \\
OntoNotes BN & BN & 65.08 & 55.58 & 59.96 & 76.5 & 57.81 & 65.86 & 59.33 & 66.12 & 62.54 & 67.0 & 59.8 & 62.8 \\
OntoNotes BC & BC & \textbf{49.13} & 30.14 & 37.36 & \textbf{55.47} & 36.56 & 44.07 & \textbf{36.33} & 50.79 & 42.36 & 47.0 & 39.2 & 41.3 \\
OntoNotes WB & WEB & \textbf{50.41} & 22.02 & \textbf{30.65} & \textbf{57.46} & 28.83 & 38.4 & 51.39 & 48.16 & 49.72 & 53.1 & 33.0 & 39.6 \\
OntoNotes TC & TC & 67.18 & 22.82 & 34.07 & 65.25 & 29.44 & 40.58 & 59.92 & 50 & 54.51 & 64.1 & 34.1 & 43.1 \\
OntoNotes MZ & MZ & 58.15 & 44.1 & 50.16 & 74.59 & 43.84 & 55.22 & 54.19 & 54.64 & 54.41 & 62.3 & 47.5 & 53.3 \\
\hline
MSM 2013 Test & TWI & 70.98 & 36.38 & 48.11 & 75.37 & 38.9 & 51.31 & 56.7 & 60.89 & 58.72 & 67.7 & 45.4 & 52.7 \\
Ritter & TWI & 75.56 & 25.19 & 37.78 & 78.29 & 29.38 & 42.73 & 59.06 & 46.67 & 52.14 & 71.0 & 33.7 & 44.2 \\
UMBC & TWI & \textbf{47.62} & \textbf{10.64} & \textbf{17.39} & 62.22 & \textbf{14.81} & \textbf{23.93} & \textbf{33.15} & \textbf{31.75} & \textbf{32.43} & 47.7 & 19.1 & 24.6 \\
\hline\hline
{\bf Standard Deviation} & & 12.46 & 16.04 & 15.73 & 15.27 & 15.56 & 15.40 & 14.24 & 15.04 & 14.18 & 14.0 & 15.5 & 15.1 \\
\hline
{\bf Macro Average} & & 58.92 & 30.82 & 38.64 & 63.00 & 35.36 & 44.13 & 49.59 & 49.17 & 48.98 & 57.2 & 38.5 & 43.9 \\
\bottomrule
\end{tabular}
\caption{\label{tab:F1Norm} P, R and F1 of NERC with different models evaluated on different testing corpora, trained on corpora normalised by size}
\end{sidewaystable*}

\setlength{\tabcolsep}{0.2em}
\begin{table*}[t]
\fontsize{9}{11}\selectfont
\centering
\begin{tabular}{c|c c c|c c c|c c c|c c c}
\toprule
&  \multicolumn{3}{c}{\bf CRFSuite} & \multicolumn{3}{c}{\bf Stanford NER} & \multicolumn{3}{c}{\bf SENNA} & \multicolumn{3}{c}{\bf Average}\\
\bf Genre & \bf P & \bf R & \bf F1 & \bf P & \bf R & \bf F1 & \bf P & \bf R & \bf F1  & \bf P & \bf R & \bf F1 \\
\hline
NW & 60.65 & 43.12 & 49.94 & 65.04 & 46.39 & 53.52 & 52.44 & 56.34 & 54.10 & 59.37 & 48.62 & 52.52 \\
BN & 60.89 & 34.57 & 40.92 & 66.30 & 42.23 & 50.99 & 49.70 & 51.48 & 50.46 & 58.96 & 42.76 & 47.46 \\
BC & 54.30 & 30.01 & 38.57 & 57.99 & 38.32 & 46.14 & 37.90 & 46.37 & 41.51 & 50.06 & 38.23 & 42.07 \\
WEB & 57.95 & 21.84 & 31.60 & 58.49 & 25.70 & 35.57 & 52.23 & 40.55 & 45.19 & 56.22 & 29.36 & 37.45 \\
CTS & 69.77 & 14.61 & 24.15 & 74.76 & 23.05 & 35.24 & 72.36 & 68.00 & 70.11 & 72.30 & 35.22 & 43.17 \\
UN & 20.00 & 0.41 & 0.81 & 10.81 & 1.65 & 2.87 & 12.59 & 7.44 & 9.35 & 14.47 & 3.17 & 4.34 \\
TC & 67.18 & 22.82 & 34.07 & 65.25 & 29.44 & 40.58 & 59.92 & 50.00 & 54.51 & 64.12 & 34.09 & 43.05 \\
MZ & 58.15 & 44.10 & 50.16 & 74.59 & 43.84 & 55.22 & 54.19 & 54.64 & 54.41 & 62.31 & 47.53 & 53.26 \\
TWI & 64.72 & 24.07 & 34.43 & 71.96 & 27.70 & 39.32 & 49.64 & 46.44 & 47.76 & 62.11 & 32.73 & 40.50 \\
\hline
\bf Macro Average & 57.07 & 26.17 & 33.85 & 60.58 & 30.92 & 39.94 & 49.00 & 46.81 & 47.49 & 55.55 & 34.63 & 40.43 \\
\bf Standard Deviation & 14.72 & 13.81 & 14.95 & 19.65 & 13.91 & 15.79 & 16.45 & 16.67 & 16.42 & 16.53 & 13.52 & 14.53 \\
\bottomrule
\end{tabular}
\caption{\label{tab:F1NormPerGenre} P, R and F1 of NERC with different models evaluated on different testing corpora, trained on corpora normalised by size, metrics macro averaged by genres}
\end{table*}

\subsubsection{Dataset Sizes and Characteristics}

Table~\ref{tab:CorpTypes} shows which genres the different corpora belong to, the number of NEs and the proportions of NE classes per corpus. Sizes of NER corpora have increased over time, from MUC to OntoNotes. 

Further, the class distribution varies between corpora: while the CoNLL corpus is very balanced and contains about equal numbers of PER, LOC and ORG NEs, other corpora are not. The least balanced corpus is the MSM 2013 Test corpus, which contains 98 LOC NEs, but 1110 PER NEs.
This makes it difficult to compare NER performance here, since performance partly depends on training data size. Since comparing NER performance as such is not the goal of this paper, we will illustrate the impact of training data size by using learning curves in the next section; illustrate NERC performance on trained corpora normalised by size in Table~\ref{tab:F1Norm}; and then only use the original training data size for subsequent experiments.

\setlength{\tabcolsep}{0.3em}
\begin{table*}[t]
\fontsize{9}{11}\selectfont
\centering
\begin{tabular}{c|c|c c c c}
\toprule
\textbf{Corpus} & \textbf{Genre} & \textbf{NEs} & \textbf{Unique NEs} & \textbf{Ratio} & \textbf{Norm Ratio} \\
\hline
MUC 7 Train & NW & 552 & 232 & 2.38 & 2.24 \\
MUC 7 Dev & NW & 572 & 238 & 2.40 & 2.14 \\
MUC 7 Test & NW & 863 & 371 & 2.33 & 1.90 \\
\hline
CoNLL Train & NW & 20038 & 7228 & 2.77 & 1.83 \\
CoNLL TestA & NW & 4223 & 2154 & 1.96 & \textbf{1.28} \\
CoNLL TestB & NW & 4937 & 2338 & 2.11 & \textbf{1.31} \\
\hline
ACE NW & NW & 3835 & 1358 & 2.82 & 2.13 \\
ACE BN & BN & 2067 & 929 & 2.22 & 1.81 \\
ACE BC & BC & 1746 & 658 & 2.65 & 1.99 \\
ACE WL & WEB & 1716 & 931 & 1.84 & \textbf{1.63} \\
ACE CTS & CTS & 2667 & 329 & 8.11 & 4.82 \\
ACE UN & UN & 668 & 374 & 1.79 & \textbf{1.60} \\
\hline
OntoNotes NW & NW & 52055 & 17748 & 2.93 & 1.77 \\
OntoNotes BN & BN & 14213 & 3808 & 3.73 & 2.58 \\
OntoNotes BC & BC & 7676 & 2314 & 3.32 & 2.47 \\
OntoNotes WB & WEB & 6080 & 2376 & 2.56 & 1.99 \\
OntoNotes TC & TC & 1430 & 717 & 1.99 & \textbf{1.66} \\
OntoNotes MZ & MZ & 8150 & 2230 & 3.65 & 3.16 \\
\hline
MSM 2013 Train & TWI & 2815 & 1817 & 1.55 & \textbf{1.41} \\
MSM 2013 Test & TWI & 1432 & 1028 & 1.39 & \textbf{1.32} \\
Ritter & TWI & 1221 & 957 & 1.28 & \textbf{1.20} \\
UMBC & TWI & 506 & 473 & 1.07 & \textbf{1.07} \\
\hline
\bf R & & & & 0.20 & 0.15 \\
\hline
\bf Mean & & & & 2.58 & 1.97 \\
\bf Median & & & & 2.35 & 1.82 \\
\bottomrule
\end{tabular}
\caption{\label{tab:TokTypeRatioNE} NE/Unique NE ratios and normalised NE/Unique NE ratios of different corpora, mean and median of those values plus R correlation of ratios with Stanford NER F1 on original corpora}
\end{table*}

\setlength{\tabcolsep}{0.3em}
\begin{table*}[t]
\fontsize{9}{11}\selectfont
\centering
\begin{tabular}{c|c|c c c c}
\toprule
\textbf{Corpus} & \textbf{Genre} & \textbf{Tokens} & \textbf{Types} & \textbf{Ratio} & \textbf{Norm Ratio} \\
\hline
MUC 7 Train & NW & 8476 & 2086 & 4.06 & 3.62 \\
MUC 7 Dev & NW & 9117 & 1722 & 5.29 & 4.79 \\
MUC 7 Test & NW & 12960 & 2895 & 4.48 & 3.80 \\
\hline
CoNLL Train & NW & 204567 & 23624 & 8.66 & \textbf{2.91} \\
CoNLL TestA & NW & 34392 & 7815 & 4.40 & \textbf{2.62} \\
CoNLL TestB & NW & 39474 & 8428 & 4.68 & \textbf{2.64} \\
\hline
ACE NW & NW & 66875 & 8725 & 7.66 & 3.40 \\
ACE BN & BN & 66534 & 7630 & 8.72 & 3.40 \\
ACE BC & BC & 52758 & 5913 & 8.92 & 4.40 \\
ACE WL & WEB & 50227 & 8529 & 5.89 & \textbf{3.12} \\
ACE CTS & CTS & 58205 & 3425 & 16.99 & 7.22 \\
ACE UN & UN & 82515 & 8480 & 9.73 & 4.49 \\
\hline
OntoNotes NW & NW & 1049713 & 42716 & 24.57 & 3.69 \\
OntoNotes BN & BN & 259347 & 16803 & 15.43 & 3.77 \\
OntoNotes BC & BC & 245545 & 13218 & 18.58 & 3.95 \\
OntoNotes WB & WEB & 205081 & 17659 & 11.61 & 3.86 \\
OntoNotes TC & TC & 110135 & 5895 & 18.68 & 6.98 \\
OntoNotes MZ & MZ & 197517 & 15412 & 12.82 & 3.68 \\
\hline
MSM 2013 Train & TWI & 56722 & 10139 & 5.59 & 3.50 \\
MSM 2013 Test & TWI & 32295 & 6474 & 4.99 & 3.66 \\
Ritter & TWI & 48864 & 10587 & 4.62 & \textbf{2.78} \\
UMBC & TWI & 7037 & 3589 & 1.96 & \textbf{1.96} \\
\hline
\bf R & & & & 0.25 & -0.07 \\
\hline
\bf Mean & & & & 9.47 & 3.83 \\
\bf Median & & & & 8.16 & 3.67 \\
\bottomrule
\end{tabular}
\caption{\label{tab:TokTypeRatio} Token/type ratios and normalised token/type ratios of different corpora, mean and median of those values plus R correlation of ratios with Stanford NER F1 on original corpora}
\end{table*}

In order to compare corpus diversity across genres, we measure NE and token/type diversity~(following e.g.~\cite{Palmer1997Statistical}). Note that types are the unique tokens, so the ratio can be understood as ratio of total tokens to unique ones. Table~\ref{tab:TokTypeRatioNE} shows the ratios between the number of NEs and the number of unique NEs per corpus, while Table~\ref{tab:TokTypeRatio} reports the token/type ratios. The lower those ratios are, the more diverse a corpus is. While token/type ratios also include tokens which are NEs, they are a good measure of broader linguistic diversity.

Aside from these metrics, there are other factors which contribute to corpus diversity, including how big a corpus is and how well sampled it is, e.g. if a corpus is only about one story, it should not be surprising to see a high token/type ratio. Therefore, by experimenting on multiple corpora, from different genres and created through different methodologies, we aim to encompass these other aspects of corpus diversity.

Since the original NE and token/type ratios do not account for corpus size, Tables~\ref{tab:TokTypeRatio} and \ref{tab:TokTypeRatioNE} present also the normalised ratios. For those, a number of tokens equivalent to those in the corpus, e.g. 7037 for UMBC (Table~\ref{tab:TokTypeRatio}) or, respectively, a number of NEs equivalent to those in the corpus (506 for UMBC) are selected (Table~\ref{tab:TokTypeRatioNE}).

An easy choice of sampling method would be to sample tokens and NEs randomly. However, this would not reflect the composition of corpora appropriately. Corpora consist of several documents, tweets or blog entries, which are likely to repeat the words or NEs since they are about one story. The difference between bigger and smaller corpora is then that bigger corpora consist of more of those documents, tweets, blog entries, interviews, etc.
Therefore, when we downsample, we take the first $n$ tokens for the token/type ratios or the first $n$ NEs for the NEs/Unique NEs ratios.

Looking at the normalised diversity metrics, the lowest NE/Unique NE ratios $<= 1.5$ (in bold, Table~\ref{tab:TokTypeRatioNE}) are observed on the Twitter and CoNLL Test corpora. Seeing this for Twitter is not surprising since one would expect noise in social media text (e.g. spelling variations or mistakes) to also have an impact on how often the same NEs are seen. Observing this in the latter, though, is less intuitive and suggests that the CoNLL corpora are well balanced in terms of stories.
Low NE/Unique ratios ($<= 1.7$) can also be observed for ACE WL, ACE UN and OntoNotes TC. Similar to social media text, content from weblogs, usenet dicussions and telephone conversations also contains a larger amount of noise compared to the traditionally-studied newswire genre, so this is not a surprising result.
Corpora bearing high NE/Unique NE ratios ($> 2.5$) are ACE CTS, OntoNotes MZ and OntoNotes BN. These results are also not surprising. The telephone conversations in ACE CTS are all about the same story, and newswire and broadcast news tend to contain longer stories (reducing variety in any fixed-size set) and are more regular due to editing.

The token/type ratios reflect similar trends (Table~\ref{tab:TokTypeRatio}). Low token/type ratios $<= 2.8$ (in bold) are observed for the Twitter corpora (Ritter and UMBC), as well as for the CoNLL Test corpus.
Token/type ratios are also low ($<= 3.2$) for CoNLL Train and ACE WL. Interestingly, ACE UN and MSM Train and Test do not have low token/type ratios although they have low NE/Unique ratios. That is, many diverse persons, locations and organisations are mentioned in those corpora, but similar context vocabulary is used. Token/type ratios are high ($>= 4.4$) for MUC7 Dev, ACE BC, ACE CTS, ACE UN and OntoNotes TC. Telephone conversations (TC) having high token/type ratios can be attributed to the high amount filler words (e.g. ``uh'', ``you know''). NE corpora are generally expected to have regular language use -- for ACE, at least, in this instance.

Furthermore, it is worth pointing out that, especially for the larger corpora (e.g. OntoNotes NW), size normalisation makes a big difference. The normalised NE/Unique NE ratios drop by almost a half compared to the un-normalised ratios, and normalised Token/Type ratios drop by up to 85\%. This strengthens our argument for size normalisation and also poses the question of low NERC performance for diverse genres being mostly due to the lack of large training corpora. This is examined in Section~\ref{RQ2}.

\setlength{\tabcolsep}{0.3em}
\begin{table*}[t]
\fontsize{9}{11}\selectfont
\centering
\begin{tabular}{c|c|c c c c}
\toprule
\textbf{Corpus} & \textbf{Genre} & \textbf{NE tokens} & \textbf{O tokens} & \textbf{Density} & \textbf{Norm Density}\\
\hline
MUC 7 Train & NW & 914 & 7562 & 0.11 & 0.11 \\
MUC 7 Dev & NW & 976 & 8141 & 0.11 & 0.10 \\
MUC 7 Test & NW & 1624 & 11336 & 0.13 & 0.13 \\
\hline
CoNLL Train & NW & 29450 & 174171 & 0.14 & 0.15 \\
CoNLL TestA & NW & 6237 & 28154 & 0.18 & 0.18 \\
CoNLL TestB & NW & 7194 & 32279 & 0.18 & 0.19 \\
\hline
ACE NW & NW & 7330 & 59545 & 0.11 & 0.11 \\
ACE BN & BN & 3555 & 62979 & 0.05 & \textbf{0.06} \\
ACE BC & BC & 3127 & 49631 & 0.06 & \textbf{0.06} \\
ACE WL & WEB & 3227 & 47000 & 0.06 & 0.08 \\
ACE CTS & TC & 3069 & 55136 & 0.05 & \textbf{0.06} \\
ACE UN & UN & 1060 & 81455 & 0.01 & \textbf{0.01} \\
\hline
OntoNotes NW & NW & 96669 & 953044 & 0.09 & 0.11 \\
OntoNotes BN & BN & 23433 & 235914 & 0.09 & 0.08 \\
OntoNotes BC & BC & 13148 & 232397 & 0.05 & 0.11 \\
OntoNotes WB & WEB & 10636 & 194445 & 0.05 & \textbf{0.06} \\
OntoNotes TC & TC & 1870 & 108265 & 0.02 & \textbf{0.01} \\
OntoNotes MZ & MZ & 15477 & 182040 & 0.08 & 0.11 \\
\hline
MSM 2013 Train & TWI & 4535 & 52187 & 0.08 & 0.07 \\
MSM 2013 Test & TWI & 2480 & 29815 & 0.08 & 0.07 \\
Ritter & TWI & 1842 & 44627 & 0.04 & \textbf{0.04} \\
UMBC & TWI & 747 & 6290 & 0.11 & 0.11 \\
\hline
\bf R & & & & 0.57 & 0.62 \\
\hline
\bf Mean & & & & 0.09 & 0.09 \\
\bf Median & & & & 0.08 & 0.09 \\
\bottomrule
\end{tabular}
\caption{\label{tab:TagDensity} Tag density and normalised tag density, the proportion of tokens with NE tags to all tokens, mean and median of those values plus R correlation of density with Stanford NER F1 on original corpora}
\end{table*}

Lastly, Table~\ref{tab:TagDensity} reports tag density (percentage of tokens tagged as part of a NE), which is another useful metric of corpus diversity that can be interpreted as the information density of a corpus. What can be observed here is that the NW corpora have the highest tag density and generally tend to have higher tag density than corpora of other genres; that is, newswire bears a lot of entities.
Corpora with especially low tag density $<= 0.06$ (in bold) are the TC corpora, Ritter, OntoNotes WB, ACE UN, ACE BN and ACE BC. As already mentioned, conversational corpora, to which ACE BC also belong, tend to have many filler words, thus it is not surprising that they have a low tag density.
There are only minor differences between the tag density and the normalised tag density, since corpus size as such does not impact tag density.

\subsection{NER Models and Features}\label{sec:models}

To avoid system-specific bias in our experiments, three widely-used supervised statistical approaches to NER are included: Stanford NER,\footnote{\small \url{http://nlp.stanford.edu/projects/project-ner.shtml}} SENNA,\footnote{\small \url{https://github.com/attardi/deepnl}} and CRFSuite.\footnote{\small \url{https://github.com/chokkan/crfsuite/blob/master/example/ner.py}}
These systems each have contrasting notable attributes.

Stanford NER~\cite{Finkel2005StanfordNER} is the most popular of the three, deployed widely in both research and commerce.
The system has been developed in terms of both generalising the underlying technology and also specific additions for certain languages.
The majority of openly-available additions to Stanford NER, in terms of models, gazetteers,\footnote{Gazetteers often play a key role in overcoming low NER recall~\cite{Mikheev99}.
These are sets of phrases that have a property in common, for example, a gazetteer of capital city names in English, or a gazetteer of military titles, would be a list of each of those things.} prefix/suffix handling and so on, have been created for newswire-style text. 
Named entity recognition and classification is modelled as a sequence labelling task with first-order conditional random fields (CRFs)~\cite{laffertyCrf}.

SENNA~\cite{Collobert2011Senna} is a more recent system for named entity extraction and other NLP tasks.
Using word representations and deep learning with deep convolutional neural networks, the general principle for SENNA is to avoid task-specific engineering while also doing well on multiple benchmarks.
The approach taken to fit these desiderata is to use representations induced from large unlabelled datasets, including LM2 (introduced in the paper itself) and Brown clusters~\cite{brown1992class,derczynski2016generalised}.
The outcome is a flexible system that is readily adaptable, given training data.
Although the system is more flexible in general, it relies on learning language models from unlabelled data, which might take a long time to gather and retrain. For the setup in \cite{Collobert2011Senna} language models are trained for seven weeks on the English Wikipedia, Reuters RCV1~\cite{Lew04} and parts of the Wall Street Journal, and results are reported over the CoNLL 2003 NER dataset. Reuters RCV1 is chosen as unlabelled data because the English CoNLL 2003 corpus is created from the Reuters RCV1 corpus. For this paper, we use the original language models distributed with SENNA and evaluate SENNA with the DeepNL framework~\cite{Attardi2015DeepNL}. As such, it is to some degree also biased towards the CoNLL 2003 benchmark data.

Finally, we use the classical NER approach from CRFsuite~\cite{okazaki2007crfsuite}, which also uses first-order CRFs.
This frames NER as a structured sequence prediction task, using features derived directly from the training text.
Unlike the other systems, no external knowledge (e.g. gazetteers and unsupervised representations) are used.
This provides a strong basic supervised system, and -- unlike Stanford NER and SENNA -- has not been tuned for any particular domain, giving potential to reveal more challenging domains without any intrinsic bias.

We use the feature extractors natively distributed with the NER frameworks. For Stanford NER we use the feature set ``chris2009'' without distributional similarity, which has been tuned for the CoNLL 2003 data. This feature was tuned to handle OOV words through word shape, i.e. capitalisation of constituent characters. The goal is to reduce feature sparsity -- the basic problem behind OOV named entities -- by reducing the complexity of word shapes for long words, while retaining word shape resolution for shorter words. In addition, word clusters, neighbouring n-grams, label sequences and quasi-Newton minima search are included.\footnote{For further details, see the official feature mnemonic list at {\tt \small https://github.com/stanfordnlp/CoreNLP/blob/master/scripts/ner/ english.conll.4class.distsim.prop}. } SENNA uses word embedding features and gazetteer features; for the training configuration see {\tt \small https://github.com/attardi/deepnl\#benchmarks}. Finally, for CRFSuite, we use the provided feature extractor without POS or chunking features, which leaves unigram and bigram word features of the mention and in a window of 2 to the left and the right of the mention, character shape, prefixes and suffixes of tokens.

These systems are compared against a simple surface form memorisation tagger.
The memorisation baseline picks the most frequent NE label for each token sequence as observed in the training corpus. 
There are two kinds of ambiguity: one is overlapping sequences, e.g. if both ``New York City'' and ``New York'' are memorised as a location. 
In that case the longest-matching sequence is labelled with the corresponding NE class.
The second, class ambiguity, occurs when the same textual label refers to different NE classes, e.g. ``Google'' could either refer to the name of a company, in which case it would be labelled as ORG, or to the company's search engine, which would be labelled as O (no NE).

\section{Experiments}\label{sec:experiments}

\subsection{RQ1: NER performance with Different Approaches}\label{RQ1}
%How does NERC performance differ for corpora between different NER approaches?

\setlength{\tabcolsep}{0.2em}
\begin{sidewaystable*}[t]
\fontsize{9}{11}\selectfont
\centering
\begin{tabular}{c|c c c|c c c|c c c|c c c|c c c|c c c}
\toprule
& \multicolumn{3}{c}{\bf Memorisation} & \multicolumn{3}{c}{\bf CRFSuite} & \multicolumn{3}{c}{\bf Stanford NER} & \multicolumn{3}{c}{\bf SENNA} & \multicolumn{3}{c}{\bf Average}\\
\bf Corpus & \bf P & \bf R & \bf F1 & \bf P & \bf R & \bf F1 & \bf P & \bf R & \bf F1 & \bf P & \bf R & \bf F1  & \bf P & \bf R & \bf F1 \\
\hline
MUC 7 Dev & 44.35 & 27.4 & 33.87 & 63.49 & 61.82 & 62.64 & 69.98 & 70.1 & 70.04 & 55.12 & 73.43 & 62.97 & 58.24 & 58.19 & 57.38\\
MUC 7 Test & 50 & 18.75 & 27.27 & 63.43 & 50.64 & 56.31 & 71.11 & 51.97 & 60.05 & 53.84 & 60.95 & 57.17 & 59.60 & 45.58 & 50.20 \\
\hline
CoNLL TestA & 61.15 & 35.67 & 45.06 & 91.51 & 88.22 & 89.84 & 93.09 & 91.39 & 92.23 & 91.73 & 93.05 & 92.39 & 84.37 & 77.08 & 79.88 \\
CoNLL TestB & 54.36 & 23.01 & 32.33 & 87.24 & 82.55 & 84.83 & 88.84 & 85.42 & 87.1 & 85.89 & 87 & 86.44 & 79.08 & 69.50 & 72.68 \\
\hline
ACE NW & 48.25 & 31.59 & 38.18 & 57.11 & 44.43 & 49.98 & 55.43 & 48.06 & 51.48 & 53.45 & 50.54 & 51.95 & 53.56 & 43.66 & 47.90 \\
ACE BN & 34.24 & 20.85 & 25.92 & 55.26 & 22.37 & 31.85 & 53.61 & 33.94 & 41.57 & 51.15 & 50.84 & 50.99 & 48.57 & 32.00 & 37.58 \\
ACE BC & 48.98 & 32.05 & 38.75 & 59.04 & 46.01 & 51.72 & 59.41 & 50.93 & 54.84 & 52.22 & 47.88 & 49.96 & 54.91 & 44.22 & 48.82 \\
ACE WL & 45.26 & 5.63 & 10.01 & 62.74 & 21.65 & 32.2 & 59.72 & 22.18 & 32.34 & 51.03 & 22.83 & 31.55 & 54.69 & 18.07 & 26.53 \\
ACE CTS & 79.73 & 17.2 & 28.3 & 80.05 & 32.04 & 45.76 & 81.89 & 39.59 & 53.38 & 75.68 & 67.22 & 71.2 & 79.34 & 39.01 & 49.66 \\
ACE UN & 12.29 & 11.93 & 12.11 & 20 & 0.41 & 0.81 & 13.51 & 2.07 & 3.58 & 22.22 & 1.65 & 3.08 & 17.01 & 4.02 & 4.90 \\
\hline
OntoNotes NW & 39.01 & 31.49 & 34.85 & 82.19 & 77.35 & 79.7 & 84.89 & 80.78 & 82.78 & 79.37 & 76.76 & 78.04 & 71.37 & 66.60 & 68.84 \\
OntoNotes BN & 18.32 & 32.98 & 23.55 & 86.51 & 80.59 & 83.44 & 88.33 & 83.42 & 85.8 & 84.69 & 83.7 & 84.2 & 69.46 & 70.17 & 69.25 \\
OntoNotes BC & 17.37 & 24.08 & 20.18 & 75.59 & 65.26 & 70.04 & 76.34 & 69.02 & 72.5 & 70.38 & 73.4 & 71.86 & 59.92 & 57.94 & 58.65 \\
OntoNotes WB & 52.61 & 29.27 & 37.62 & 64.73 & 45.52 & 53.45 & 68.62 & 54.13 & 60.52 & 63.94 & 61.3 & 62.59 & 62.48 & 47.56 & 53.55 \\
OntoNotes TC & 6.48 & 16.55 & 9.32 & 65.26 & 32.4 & 43.31 & 68.82 & 44.6 & 54.12 & 73.45 & 57.84 & 64.72 & 53.50 & 37.85 & 42.87 \\
OntoNotes MZ & 44.56 & 31.12 & 36.64 & 79.87 & 74.27 & 76.97 & 82.07 & 79.32 & 80.67 & 74.42 & 76.23 & 75.31 & 70.23 & 65.24 & 67.40 \\
\hline
MSM 2013 Test & 20.51 & 7.84 & 11.35 & 83.08 & 56.91 & 67.55 & 83.64 & 60.68 & 70.34 & 70.89 & 70.74 & 70.81 & 64.53 & 49.04 & 55.01 \\
Ritter & 50.81 & 15.11 & 23.29 & 76.36 & 31.11 & 44.21 & 80.57 & 34.81 & 48.62 & 67.43 & 43.46 & 52.85 & 68.79 & 31.12 & 42.24 \\
UMBC & 76.92 & 5.29 & 9.9 & 47.62 & 10.64 & 17.39 & 62.22 & 14.81 & 23.93 & 33.15 & 31.75 & 32.43 & 54.98 & 15.62 & 20.91\\
\hline\hline
{\bf Standard Deviation} & 20.23 & 9.78 & 11.44 & 17.11 & 25.62 & 23.84 & 18.28 & 25.27 & 23.14 & 17.96 & 23.14 & 21.63 & 18.40 & 20.95 & 20.01 \\
\hline
{\bf Macro Average} & 42.38 & 21.99 & 26.24 & 68.48 & 48.64 & 54.84 & 70.64 & 53.54 & 59.26 & 63.69 & 59.50 & 60.55 & 61.30 & 45.92 & 50.22 \\
\bottomrule
\end{tabular}
\caption{\label{tab:F1} P, R and F1 of NERC with different models trained on original corpora}
\end{sidewaystable*}

\setlength{\tabcolsep}{0.2em}
\begin{sidewaystable*}[t] % sidewaystable* or table*
\fontsize{8}{10}\selectfont
\centering
\begin{tabular}{c|c c c|c c c|c c c|c c c|c c c}
\toprule
& \multicolumn{3}{c}{\bf Memorisation} & \multicolumn{3}{c}{\bf CRFSuite} & \multicolumn{3}{c}{\bf Stanford NER} & \multicolumn{3}{c}{\bf SENNA} & \multicolumn{3}{c}{\bf Average}\\
\bf Corpus & \bf PER & \bf LOC & \bf ORG & \bf PER & \bf LOC & \bf ORG & \bf PER & \bf LOC & \bf ORG & \bf PER & \bf LOC & \bf ORG  & \bf PER & \bf LOC & \bf ORG \\
\hline
MUC 7 Dev & 3.67 & 28.3 & 43.87 & 58.54 & 63.83 & 63.03 & 57.61 & 75.44 & 70.44 & 60.83 & 62.99 & 63.8 & 45.16 & 57.64 & 60.29  \\
MUC 7 Test & 16.77 & 51.15 & 9.56 & 52.87 & 59.79 & 55.33 & 61.09 & 72.8 & 50.42 & 79.08 & 61.38 & 47.4 & 52.45 & 61.28 & 40.68 \\
\hline
CoNLL TestA & 34.88 & 49.11 & 55.66 & 92 & 91.92 & 84 & 94.31 & 93.85 & 87.09 & 95.14 & 94.65 & 85.55 & 79.08 & 82.38 & 78.08 \\
CoNLL TestB & 12.8 & 40.52 & 41.71 & 87.45 & 87.09 & 79.78 & 90.47 & 89.21 & 81.45 & 91.42 & 89.35 & 78.78 & 70.54 & 76.54 & 70.43 \\
\hline
ACE NW & 0 & 52.12 & 0 & 37.47 & 57.35 & 33.89 & 39.92 & 58.46 & 38.84 & 49.72 & 57.89 & 30.99 & 31.78 & 56.46 & 25.93 \\
ACE BN & 13.37 & 41.26 & 18 & 35.58 & 35.36 & 14.81 & 45.86 & 41.56 & 31.09 & 55.74 & 55.02 & 26.73 & 37.64 & 43.30 & 22.66 \\
ACE BC & 0 & 62.64 & 0 & 44.32 & 64.07 & 23.19 & 49.34 & 65.24 & 30.77 & 44.98 & 61.81 & 18.92 & 34.66 & 63.44 & 18.22 \\
ACE WL & 1.44 & 33.06 & 0 & 39.01 & 37.5 & 10.97 & 37.41 & 42.91 & 7.96 & 37.75 & 40.15 & 5.96 & 28.90 & 38.41 & 6.22 \\
ACE CTS & 18.62 & 63.77 & 0 & 46.28 & 46.64 & 0 & 52.67 & 60.56 & 0 & 75.31 & 47.35 & 26.09 & 48.22 & 54.58 & 6.52 \\
ACE UN & 0 & 14.92 & 7.55 & 3.08 & 0 & 0 & 0 & 6.17 & 0 & 5.48 & 2.78 & 0 & 2.14 & 5.97 & 1.89 \\
\hline
OntoNotes NW & 20.85 & 55.67 & 14.85 & 84.75 & 82.39 & 74.82 & 86.39 & 85.62 & 78.47 & 80.82 & 85.71 & 69.08 & 68.20 & 77.35 & 59.31 \\
OntoNotes BN & 15.37 & 27.52 & 12.59 & 88.67 & 86.4 & 66.64 & 90.98 & 88.88 & 68.78 & 90.06 & 87.46 & 66.24 & 71.27 & 72.57 & 53.56 \\
OntoNotes BC & 10.92 & 24.89 & 10.9 & 69.13 & 79.28 & 52.26 & 75.54 & 79.04 & 54.48 & 74.13 & 81.02 & 50.98 & 57.43 & 66.06 & 42.16 \\
OntoNotes WB & 26.2 & 54.86 & 5.71 & 50.17 & 67.76 & 22.33 & 60.28 & 72.05 & 30.21 & 66.63 & 73.15 & 31.03 & 50.82 & 66.96 & 22.32 \\
OntoNotes TC & 19.67 & 7.71 & 0 & 40.68 & 50.38 & 8.16 & 54.24 & 58.43 & 18.87 & 67.06 & 68.39 & 7.84 & 45.41 & 46.23 & 8.72 \\
OntoNotes MZ & 21.39 & 58.76 & 4.55 & 83.49 & 80.44 & 52.86 & 86.44 & 84.65 & 56.88 & 83.92 & 78.06 & 48.25 & 68.81 & 75.48 & 40.64 \\
\hline
MSM 2013 Test & 4.62 & 44.71 & 26.14 & 75.9 & 43.75 & 24.14 & 78.46 & 42.77 & 31.37 & 81.04 & 45.42 & 28.12 & 60.01 & 44.16 & 27.44 \\
Ritter & 15.54 & 35.11 & 20.13 & 43.19 & 54.08 & 33.54 & 48.65 & 57 & 37.97 & 63.64 & 61.47 & 22.22 & 42.76 & 51.92 & 28.47 \\
UMBC & 5.97 & 22.22 & 0 & 8.33 & 32.61 & 6.06 & 25.32 & 39.08 & 2.94 & 33.12 & 52.1 & 6.38 & 18.19 & 36.50 & 3.85 \\
\hline\hline
{\bf Standard Deviation} & 9.83 & 16.42 & 16.67 & 26.12 & 23.40 & 28.02 & 25.05 & 22.39 & 28.03 & 23.02 & 21.61 & 26.00 & 21.00 & 20.95 & 24.68 \\
\hline
{\bf Macro Average} & 12.74 & 40.44 & 14.27 & 54.78 & 58.98 & 37.15 & 59.74 & 63.88 & 40.95 & 65.05 & 63.48 & 37.60 & 48.08 & 56.69 & 32.49 \\
\bottomrule
\end{tabular}
\caption{\label{tab:F1PerType} F1 per NE type with different models trained on original corpora}
\end{sidewaystable*}

\begin{figure}
\centering
\includegraphics[width=1\textwidth]{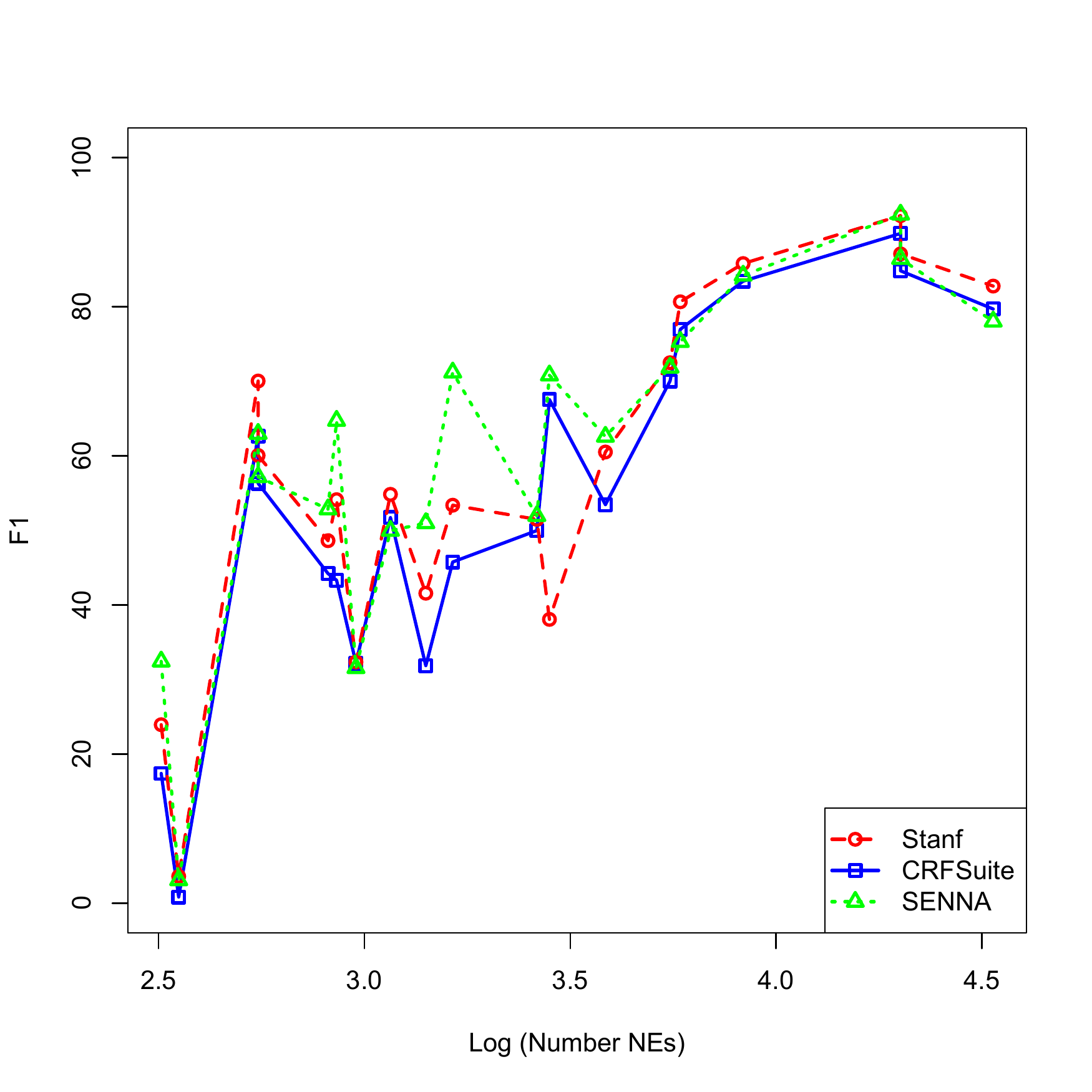}
\caption{F1 of different NER methods with respect to training corpus size, measured in log of number of NEs}\label{fig:F1CorpSize}
\end{figure}

Our first research question is how NERC performance differs for corpora between approaches.
In order to answer this, Precision (P), Recall (R) and F1 metrics are reported on size-normalised corpora (Table~\ref{tab:F1Norm}) and original corpora (Tables~\ref{tab:F1} and~\ref{tab:F1PerType}).
The reason for size normalisation is to make results comparable across corpora.
For size normalisation, the training corpora are downsampled to include the same number of NEs as the smallest corpus, UMBC. For that, sentences are selected from the beginning of the train part of the corpora so that they include the same number of NEs as UMBC. Other ways of downsampling the corpora would be to select the first $n$ sentences or the first $n$ tokens, where $n$ is the number of sentences in the smallest corpus. The reason that the number of NEs, which represent the number of positive training examples, is chosen for downsampling the corpora is that the number of positive training examples have a much bigger impact on learning than the number of negative training examples. For instance,~\cite{LearningFromLittle}, among others, study topic classification performance for small corpora and sample from the Reuters corpus. They find that adding more negative training data gives little to no improvement, whereas adding positive examples drastically improves performance.

Table~\ref{tab:F1Norm} shows results with size normalised precision (P), recall (R), and F1-Score (F1). The five lowest P, R and F1 values per method (CRFSuite, Stanford NER, SENNA) are in bold to highlight underperformers. Results for all corpora are summed with macro average.

Comparing the different methods, the highest F1 results are achieved with SENNA, followed by Stanford NER and CRFSuite. SENNA has a balanced P and R, which can be explained by the use of word embeddings as features, which help with the unseen word problem. For Stanford NER as well as CRFSuite, which do not make use of embeddings, recall is about half of precision. These findings are in line with other work reporting the usefulness of word embeddings and deep learning for a variety of NLP tasks and domains~\cite{conf/icml/SocherLNM11,conf/icml/GlorotBB11,bengio2012deep}.
With respect to individual corpora, the ones where SENNA outperforms other methods by a large margin ($>=$ 13 points in F1) are CoNLL Test A, ACE CTS and OntoNotes TC. The first success can be attributed to being from the same the domain SENNA was originally tuned for. The second is more unexpected and could be due to those corpora containing a disproportional amount of PER and LOC NEs (which are easier to tag correctly) compared to ORG NEs, as can be seen in Table~\ref{tab:F1PerType}, where F1 of NERC methods is reported on the original training data. 

Our analysis of CRFSuite here is that it is less tuned for NW corpora and might therefore have a more balanced performance across genres does not hold. Results with CRFSuite for every corpus are worse than the results for that corpus with Stanford NER, which is also CRF-based.

To summarise, our findings are:
\begin{itemize}[noitemsep]
\item{F1 is highest with SENNA, followed by Stanford NER and CRFSuite}
\item{SENNA outperforms other methods by a large margin (e.g. $>=$ 13 points in F1) for CoNLL Test A, ACE CTS and OntoNotes TC}
\item{Our hypothesis that CRFSuite is less tuned for NW corpora and will therefore have a more balanced performance across genres does not hold, as results for CRFSuite for every corpus are worse than with Stanford NER}
\end{itemize}

\subsection{RQ2: NER performance in Different Genres}\label{RQ2}

Our second research question is whether existing NER approaches generalise well over corpora in different genres.
To do this we study again Precision (P), Recall (R) and F1 metrics on size-normalised corpora (Table~\ref{tab:F1Norm}), on original corpora (Tables~\ref{tab:F1} and~\ref{tab:F1PerType}), and we further test performance per genre in a separate table (Table~\ref{tab:F1NormPerGenre}).

F1 scores over size-normalised corpora vary widely (Table~\ref{tab:F1Norm}). For example, the SENNA scores range from 9.35\% F1 (ACE UN) to 71.48\% (CoNLL Test A). Lowest results are consistently observed for the ACE subcorpora, UMBC, and OntoNotes BC and WB. The ACE corpora are large and so may be more prone to non-uniformities emerging during downsampling; they also have special rules for some kinds of organisation which can skew results (as described in Section~\ref{sec:corpora}). The highest results are on the CoNLL Test A corpus, OntoNotes BN and MUC 7 Dev. This moderately supports our hypothesis that NER systems perform better on NW than on other genres, probably due to extra fitting from many researchers using them as benchmarks for tuning their approaches. Looking at the Twitter (TWI) corpora present the most challenge due to increased diversity, the trends are unstable. Although results for UMBC are among the lowest, results for MSM 2013 and Ritter are in the same range or even higher than those on NW datasets. This begs the question whether low results for Twitter corpora reported previously were due to the lack of sufficient in-genre training data.

Comparing results on normalised to non-normalised data, Twitter results are lower than those for most OntoNotes corpora and CoNLL test corpora, mostly due to low recall. 
Other difficult corpora having low performance are ACE UN and WEB corpora. We further explicitly examine results on size normalised corpora grouped by corpus type, shown in Table~\ref{tab:F1NormPerGenre}. 
It becomes clear that, on average, newswire corpora and OntoNotes MZ are the easiest corpora and ACE UN, WEB and TWI are harder.
This confirms our hypothesis that social media and Web corpora are challenging for NERC.

The CoNLL results, on the other hand, are the highest across all corpora irrespective of the NERC method. What is very interesting to see is that they are much higher than the results on the biggest training corpus, OntoNotes NW. For instance, SENNA has an F1 of 78.04 on OntoNotes, compared to an F1 of 92.39 and 86.44 for CoNLL Test A and Test B respectively. So even though OntoNotes NW is more than twice the size of CoNLL in terms of NEs (see Table~\ref{tab:TokTypeRatioNE}), NERC performance is much higher on CoNLL. NERC performance with respect to training corpus size is represented in Figure~\ref{fig:F1CorpSize}. The latter figure confirms that although there is some correlation between corpus size and F1, the variance between results on comparably sized corpora is big. This strengthens our argument that there is a need for experimental studies, such as those reported below, to find out what, apart from corpus size, impacts NERC performance.

Another set of results presented in Table~\ref{tab:F1} are those of the simple NERC memorisation baseline. It can be observed that corpora with a low F1 for NERC methods, such as UMBC and ACE UN, also have a low memorisation performance. Memorisation is discussed in more depth in Section~\ref{sec:RQ5}.

% For corpora with high NERC performance this holds again, but not for all corpora, e.g. for CoNLL Test A and OntoNotes NW, both memorisation performance and NERC performance is high, but memorisation performance for OntoNotes BC is only average.

When NERC results are compared to the corpus diversity statistics, i.e. NE/Unique NE ratios (Table~\ref{tab:TokTypeRatioNE}), token/type ratios (Table~\ref{tab:TokTypeRatio}), and tag density (Table~\ref{tab:TagDensity}), the strongest predictor for F1 is tag density, as can be evidenced by the R correlation values between the ratios and F1 scores with the Stanford NER system, shown in the respective tables.

There is a positive correlation between high F1 and high tag density (R of 0.57 and R of 0.62 with normalised tag density), a weak positive correlation for NE/unique ratios (R of 0.20 and R of 0.15 for normalised ratio), whereas for token/type ratios, no such clear correlation can be observed (R of 0.25 and R of -0.07 for normalised ratio).

However, tag density is also not an absolute predictor for NERC performance. 
While NW corpora have both high NERC performance and high tag density, this high density is not necessarily an indicator of high performance.
For example, systems might not find high tag density corpora of other genres necessarily so easy.

One factor that can explain the difference in genre performance between e.g. newswire and social media is entity drift -- the change in observed entity terms over time.
In this case, it is evident from the differing surface forms and contexts for a given entity class.
For example, the concept of ``location" that NER systems try to learn might be frequently represented in English newswire from 1991 with terms like {\em Iraq} or {\em Kuwait}, but more with {\em Atlanta}, {\em Bosnia} and {\em Kabul} in the same language and genre from 1996.
Informally, drift on Twitter is often characterised as both high-frequency and high-magnitude; that is, the changes are both rapid and correspond to a large amount of surface form occurrences (e.g.~\cite{fromreide2014crowdsourcing,derczynski2015usfd}).

\begin{table}[t]
\fontsize{9}{11}\selectfont
\centering
\begin{tabular}{ll|rrrr} 
&& \multicolumn{4}{c}{\textbf{Target}} \\
& \bf Coverage & Ritter & UMBC & MSM & W-NUT \\
\hline
\multirow{4}{*}{\textbf{Seed}}
&Ritter (2010)   & - &  9.3\% & 14.3\% & 11.1\% \\
&UMBC (2011)     &  5.0\% & - &  6.1\% &  3.0\% \\
&MSM 2013 Train  & 16.2\% & 12.1\% & - & 10.4\% \\
&W-NUT (2015)    &  5.3\% &  3.2\% &  2.6\% & - \\
\bottomrule
\end{tabular}
\caption{Entity surface form occurrence overlap between Twitter corpora.}
\label{tab:surface-drift-nwire}
\end{table}

\begin{table}[t]
\fontsize{9}{11}\selectfont
\centering
\begin{tabular}{ll|rrrr}
&& \multicolumn{4}{c}{\textbf{Target}} \\
& \bf Coverage & MUC & CoNLL & ACE & OntoNotes \\
\hline
\multirow{4}{*}{\textbf{Seed}}
&MUC 7 (1996)           & - &  6.8\% &  9.1\% &  5.7\% \\
&CoNLL Train (1996-7)   & 33.0\% & - & 48.7\% & 20.1\% \\
&ACE NW (2000-3)        & 14.8\% & 18.6\% & - & 10.4\% \\
&OntoNotes NW (2007-14) & 50.5\% & 35.4\% & 57.3\% & - \\
\bottomrule
\end{tabular}
\caption{Entity surface form occurrence overlap between news corpora.}
\label{tab:surface-drift-tweet}
\end{table}

We examined the impact of drift in newswire and Twitter corpora, taking datasets based in different timeframes.
The goal is to gauge how much diversity is due to new entities appearing over time.
To do this, we used just the surface lexicalisations of entities as the entity representation.
The overlap of surface forms was measured across different corpora of the same genre and language.
We used an additional corpus based on recent data -- that from the W-NUT 2015 challenge~\cite{baldwin2015shared}.
This is measured in terms of occurrences, rather than distinct surface forms, so that the magnitude of the drift is shown instead of having skew in results from the the noisy long tail.
Results are given in Table~\ref{tab:surface-drift-nwire} for newswire and Table~\ref{tab:surface-drift-tweet} for Twitter corpora.

It is evident that the within-class commonalities in surface forms are much higher in newswire than in Twitter.
That is to say, observations of entity texts in one newswire corpus are more helpful in labelling other newswire corpora, than if the same technique is used to label other twitter corpora.

This indicates that drift is lower in newswire than in tweets.
Certainly, the proportion of entity mentions in most recent corpora (the rightmost-columns) are consistently low compared to entity forms available in earlier data.
These reflect the raised OOV and drift rates found in previous work~\cite{fromreide2014crowdsourcing,derczynski2016broad}.
Another explanation is that there is higher noise in variation, and that the drift is not longitudinal, but rather general.
This is partially addressed by RQ3, which we will address next, in Section~\ref{sec:models:sec:RQ3}.

%In summary, observations in this section are that NERC approaches perform particularly well on the CoNLL corpus, and, in general, better on NW corpora than on most other genres. However, normalising corpora by size results in more noisy data such as TWI and WEB data achieving similar results to NW corpora. Therefore, one conclusion is that increasing the amount of available in-domain training data will likely result in improved NERC performance. Corpus size, however, is not an absolute predictor, since NW corpora larger than the CoNLL dataset still achieve lower NERC performance. A high tag density is a good, but also not absolute, predictor for high F1. What we found to be a good predictor for a high F1 is a high memorisation performance. Inspired by those findings, the next section will take a closer look at the impact of seen and unseen NEs on NER performance.

To summarise, our findings are:
% Removed memorisation comments since R1 doesn't want us to talk about it here, but rather later
\begin{itemize}[noitemsep]
\item{Overall, F1 scores vary widely across corpora.}
\item{Trends can be marked in some genres. On average, newswire corpora and OntoNotes MZ are the easiest corpora and ACE UN, WEB and TWI are the hardest corpora for NER methods to reach good performance on.}
\item{Normalising corpora by size results in more noisy data such as TWI and WEB data achieving similar results to NW corpora.}
\item{Increasing the amount of available in-domain training data will likely result in improved NERC performance.}
\item{There is a strong positive correlation between high F1 and high tag density, a weak positive correlation for NE/unique ratios and no clear correlation between token/type ratios and F1}
\item{Temporal NE drift is lower in newswire than in tweets}
\end{itemize}

The next section will take a closer look at the impact of seen and unseen NEs on NER performance.

\setlength{\tabcolsep}{0.3em}
\begin{table}%[!htbp]
\fontsize{9}{11}\selectfont
\centering
\begin{tabular}{c c|c c|c}
\toprule
\bf Corpus & \bf Genre & \bf Unseen & \bf Seen & \bf Percentage Unseen \\
\hline
MUC 7 Dev & NW & 348 & 224 & 0.608 \\
MUC 7 Test & NW &  621 & 241 & 0.720 \\
\hline
CoNLL TestA & NW  & 1485 & 2744 & 0.351 \\
CoNLL TestB & NW  & 2453 & 2496 & 0.496 \\
\hline
ACE NW & NW & 549 & 662 & 0.453 \\
ACE BN & BN & 365 & 292 & 0.556 \\
ACE BC & BC & 246 & 343 & 0.418 \\
ACE WL & WEB & 650 & 112 & 0.853 \\
ACE CTS & TC & 618 & 410 & 0.601 \\
ACE UN & UN & 274 & 40 & 0.873 \\
\hline
OntoNotes NW & NW & 8350 & 10029 & 0.454 \\
OntoNotes BN & BN & 2427 & 3470 & 0.412 \\
OntoNotes BC & BC & 1147 & 1003 & 0.533 \\
OntoNotes WB & WEB & 1390 & 840 & 0.623 \\
OntoNotes TC & TC & 486 & 88 & 0.847 \\
OntoNotes MZ & MZ & 1185 & 1112 & 0.516 \\
\hline
MSM 2013 Test & TWI & 992 & 440 & 0.693 \\
Ritter & TWI & 302 & 103 & 0.746 \\
UMBC & TWI & 176 & 13 & 0.931 \\
\bottomrule
\end{tabular}
\caption{\label{tab:SeenUnseen} Proportion of unseen entities in different test corpora}
\end{table}

\setlength{\tabcolsep}{0.2em}
\begin{table*}%[!htbp] % works as sidewaystable too, but for submission, we'll need to decrease the font size
\fontsize{8}{10}\selectfont
\centering
\begin{tabular}{c c|c c c|c c c|c c c|c c c}
\toprule
& &  \multicolumn{3}{c}{\bf CRFSuite} & \multicolumn{3}{c}{\bf Stanf} & \multicolumn{3}{c}{\bf SENNA} & \multicolumn{3}{c}{\bf All} \\
&  & \bf P & \bf R & \bf F1 & \bf P & \bf R & \bf F1 & \bf P & \bf R & \bf F1 & \bf P & \bf \bf R & \bf F1 \\
\hline\hline
MUC 7 Dev & Seen & 96.00 & 85.71 & 90.57 & 98.65 & 98.21 & 98.43 & 90.48 & 93.3 & 91.87 & 95.04 & 92.41 & 93.62 \\
& Unseen & 65.22 & 47.41 & 54.91 & 75.31 & 52.59 & 61.93 & 59.54 & 60.06 & 59.8 & 66.69 & 53.35 & 58.88 \\
\hline
MUC 7 Test & Seen & 87.02 & 74.79 & 80.44 & 97.81 & 92.53 & 95.10 & 82.81 & 87.6 & 85.14 & 89.21 & 84.97 & 86.89 \\
& Unseen & 59.95 & 41.22 & 48.85 & 61.93 & 37.2 & 46.48 & 51.57 & 50.24 & 50.9 & 57.82 & 42.89 & 48.74 \\
\hline\hline
CoNLL TestA & Seen & 96.79 & 94.46 & 95.61 & 97.87 & 97.34 & 97.61 & 97.34 & 96.55 & 96.94 & 97.33 & 96.12 & 96.72 \\
& Unseen & 86.34 & 79.19 & 82.61 & 87.85 & 81.82 & 84.73 & 95.32 & 92.43 & 93.85 & 89.84 & 84.48 & 87.06 \\
\hline
CoNLL TestB & Seen & 93.70 & 89.98 & 91.80 & 96.07 & 94.07 & 95.06 & 94.3 & 91.77 & 93.02 & 94.69 & 91.94 & 93.29 \\
& Unseen & 85.71 & 76.76 & 80.99 & 86.76 & 79.05 & 82.72 & 91.69 & 87.32 & 89.46 & 88.05 & 81.04 & 84.39 \\
\hline\hline
ACE NW & Seen & 97.28 & 64.80 & 77.79 & 96.05 & 69.79 & 80.84 & 93.3 & 63.14 & 75.32 & 95.54 & 65.91 & 77.98 \\
& Unseen & 57.48 & 22.40 & 32.24 & 56.05 & 25.32 & 34.88 & 63.49 & 36.43 & 46.3 & 59.01 & 28.05 & 37.81 \\
\hline
ACE BN & Seen & 93.65 & 40.41 & 56.46 & 94.29 & 67.81 & 78.88 & 91.32 & 68.49 & 78.28 & 93.09 & 58.90 & 71.21 \\
& Unseen & 66.04 & 9.59 & 16.75 & 47.44 & 10.14 & 16.7 & 71.57 & 40 & 51.32 & 61.68 & 19.91 & 28.26 \\
\hline
ACE BC & Seen & 90.76 & 65.89 & 76.35 & 91.01 & 70.85 & 79.67 & 88.11 & 62.68 & 73.25 & 89.96 & 66.47 & 76.42 \\
& Unseen & 62.82 & 19.92 & 30.25 & 62.89 & 24.8 & 35.57 & 57.5 & 28.05 & 37.7 & 61.07 & 24.26 & 34.51 \\
\hline
ACE WL & Seen & 89.47 & 60.71 & 72.34 & 96.67 & 77.68 & 86.14 & 91.49 & 38.39 & 54.09 & 92.54 & 58.93 & 70.86 \\
& Unseen & 75.76 & 15.38 & 25.58 & 61.03 & 12.77 & 21.12 & 62.21 & 20.77 & 31.14 & 66.33 & 16.31 & 25.95 \\
\hline
ACE CTS & Seen & 97.38 & 45.37 & 61.90 & 98.48 & 63.17 & 76.97 & 96.35 & 64.39 & 77.19 & 97.40 & 57.64 & 72.02 \\
& Unseen & 95.42 & 23.66 & 37.92 & 92.55 & 24.11 & 38.25 & 96.43 & 69.9 & 81.05 & 94.80 & 39.22 & 52.41 \\
\hline
ACE UN & Seen & 0.00 & 0.00 & 0.00 & 0 & 0 & 0 & 100 & 1.53 & 3.02 & 33.33 & 0.51 & 1.01 \\
& Unseen & 100.00 & 0.51 & 1.02 & 62.5 & 1.82 & 3.55 & 0 & 0 & 0 & 54.17 & 0.78 & 1.52 \\
\hline\hline
OntoNotes NW & Seen & 95.18 & 90.44 & 92.75 & 96.88 & 93.98 & 95.4 & 73.12 & 65.76 & 69.24 & 88.39 & 83.39 & 85.80 \\
& Unseen & 73.43 & 63.00 & 67.81 & 76.17 & 65.8 & 70.6 & 96.88 & 93.98 & 95.4 & 82.16 & 74.26 & 77.94 \\
\hline
OntoNotes BN & Seen & 95.60 & 90.86 & 93.17 & 96.75 & 94.5 & 95.61 & 81.76 & 73.34 & 77.32 & 91.37 & 86.23 & 88.70 \\
& Unseen & 82.67 & 67.24 & 74.16 & 83.45 & 68.97 & 75.52 & 96.75 & 94.5 & 95.61 & 87.62 & 76.90 & 81.76 \\
\hline
OntoNotes BC & Seen & 95.29 & 88.83 & 91.95 & 93.85 & 88.24 & 90.96 & 64.27 & 59.11 & 61.58 & 84.47 & 78.73 & 81.50 \\
& Unseen & 70.91 & 47.60 & 56.96 & 74.82 & 55.19 & 63.52 & 93.85 & 88.24 & 90.96 & 79.86 & 63.68 & 70.48 \\
\hline
OntoNotes WB & Seen & 91.96 & 81.57 & 86.45 & 94.01 & 89.64 & 91.77 & 63.75 & 47.73 & 54.59 & 83.24 & 72.98 & 77.60 \\
& Unseen & 58.97 & 26.49 & 36.56 & 64.86 & 34.39 & 44.95 & 94.01 & 89.64 & 91.77 & 72.61 & 50.17 & 57.76 \\
\hline
OntoNotes TC & Seen & 94.03 & 56.25 & 70.39 & 94.81 & 82.95 & 88.48 & 80.2 & 51.73 & 62.89 & 89.68 & 63.64 & 73.92 \\
& Unseen & 70.79 & 27.27 & 39.38 & 74.8 & 37.86 & 50.27 & 94.81 & 82.95 & 88.48 & 80.13 & 49.36 & 59.38 \\
\hline
OntoNotes MZ & Seen & 95.24 & 88.89 & 91.95 & 99.09 & 97.75 & 98.42 & 71.31 & 62.86 & 66.82 & 88.55 & 83.17 & 85.73 \\
& Unseen & 75.44 & 57.95 & 65.55 & 80.23 & 64.05 & 71.23 & 99.09 & 97.75 & 98.42 & 84.92 & 73.25 & 78.40 \\
\hline\hline
MSM 2013 Test & Seen & 92.40 & 69.09 & 79.06 & 91.73 & 78.18 & 84.42 & 84.22 & 69.96 & 76.43 & 89.45 & 72.41 & 79.97 \\
& Unseen & 87.21 & 52.22 & 65.32 & 87.08 & 54.33 & 66.91 & 91.73 & 78.18 & 84.42 & 88.67 & 61.58 & 72.22 \\
\hline
Ritter & Seen & 100.00 & 65.05 & 78.82 & 98.8 & 79.61 & 88.17 & 100 & 68.93 & 81.61 & 99.60 & 71.20 & 82.87 \\
& Unseen & 79.73 & 19.54 & 31.38 & 76.62 & 19.54 & 31.13 & 78.17 & 36.75 & 50 & 78.17 & 25.28 & 37.50 \\
\hline
UMBC & Seen & 100.00 & 23.08 & 37.50 & 100 & 53.85 & 70 & 90 & 69.23 & 78.26 & 96.67 & 48.72 & 61.92 \\
& Unseen & 59.38 & 10.86 & 18.36 & 66.67 & 12.5 & 21.05 & 52.78 & 32.39 & 40.14 & 59.61 & 18.58 & 26.52 \\
\hline\hline
{\bf Macro Average} & Seen & 89.57 & 67.17 & 75.02  & 91.20 & 78.43 & 83.79 & 92.99 & 74.38	& 79.85 & 91.25 & 73.32 & 79.55 \\
& Unseen & 74.38 & 37.27 & 45.61 & 72.58 & 40.12 & 48.48 & 73.63 & 51.91 & 58.08 & 73.53	& 43.10 & 50.72  \\
\bottomrule
\end{tabular}
\caption{\label{tab:F1Singletons} P, R and F1 of NERC with different models of unseen and seen NEs}
\end{table*}

\subsection{RQ3: Impact of NE Diversity}\label{sec:models:sec:RQ3}

\textit{Unseen} NEs are those with surface forms present only in the test, but not training data, whereas \textit{seen} NEs are those also encountered in the training data. As discussed previously, the ratio between those two measures is an indicator of corpus NE diversity.   

Table~\ref{tab:SeenUnseen} shows how the number of unseen NEs per test corpus relates to the total number of NEs per corpus. 
The proportion of unseen forms varies widely by corpus, ranging from 0.351 (ACE NW) to 0.931 (UMBC).
As expected there is a correlation between corpus size and percentage of unseen NEs, i.e. smaller corpora such as MUC and UMBC tend to contain a larger proportion of unseen NEs than bigger corpora such as ACE NW. 
In addition, similar to the token/type ratios listed in Table~\ref{tab:TokTypeRatio}, we observe that TWI and WEB corpora have a higher proportion of unseen entities.

As can be seen from Table~\ref{tab:F1}, corpora with a low percentage of unseen NEs (e.g. CoNLL Test A and OntoNotes NW) tend to have high NERC performance, whereas corpora with high percentage of unseen NEs (e.g. UMBC) tend to have low NERC performance. This suggests that systems struggle to recognise and classify unseen NEs correctly. 

To check this seen/unseen performance split, next we examine NERC performance for unseen and seen NEs separately; results are given in Table~\ref{tab:F1Singletons}.
The ``All"  column group represents an averaged performance result.
What becomes clear from the macro averages\footnote{Note
that the performance over unseen and seen entities in Table~\ref{tab:F1Singletons} does not add up to the performance reported in Table~\ref{tab:F1} because performance in Table~\ref{tab:F1Singletons} is only reported on positive test samples.}
is that F1 on unseen NEs is significantly lower than F1 on seen NEs for all three NERC approaches. This is mostly due to recall on unseen NEs being lower than that on seen NEs, and suggests some memorisation and poor generalisation in existing systems. In particular, Stanford NER and CRFSuite have almost 50\% lower recall on unseen NEs compared to seen NEs.
One outlier is ACE UN, for which the average seen F1 is 1.01 and the average unseen F1 is 1.52, though both are miniscule and the different negligible.
%Since the ACE test corpus is very small, containing 314 NEs in total result could be due to chance. IA: better not mention that otherwise they'll ask for significance values next.

Of the three approaches, SENNA exhibits the narrowest F1 difference between seen and unseen NEs. 
In fact it performs below Stanford NER for seen NEs on many corpora.
This may be because SENNA has but a few features, based on word embeddings, which reduces feature sparsity; intuitively, the simplicity of the representation is likely to help with unseen NEs, at the cost of slightly reduced performance on seen NEs through slower fitting.
Although SENNA appears to be better at generalising than Stanford NER and our CRFSuite baseline, the difference between its performance on seen NEs and unseen NEs is still noticeable.
This is 21.77 for SENNA (macro average), whereas it is 29.41 for CRFSuite and 35.68 for Stanford NER.

The fact that performance over unseen entities is significantly lower than on seen NEs partly explains what we observed in the previous section; i.e., that corpora with a high proportion of unseen entities, such as the ACE WL corpus, are harder to label than corpora of a similar size from other genres, such as the ACE BC corpus (e.g. systems reach F1 of $\sim$30 compared to $\sim$50; Table~\ref{tab:F1}).

However, even though performance on seen NEs is higher than on unseen, there is also a difference between seen NEs in corpora of different sizes and genres.
For instance, performance on seen NEs in ACE WL is 70.86 (averaged over the three different approaches), whereas performance on seen NEs in the less-diverse ACE BC corpus is higher at 76.42; the less diverse data is, on average, easier to tag.
Interestingly, average F1 on seen NEs in the Twitter corpora (MSM and Ritter) is around 80, whereas average F1 on the ACE corpora, which are of similar size, is lower, at around 70. 
%Amongst the three smallest corpora (UMBC, MUC Dev and MUC Test), averaged performance on seen NEs in the UMBC corpora is significantly lower (61.92 vs. 93.62 and 86.89 respectively). 

%To summarise, NE diversity explains a large part of the F1 differences of NER approaches between genres, but not all of it.
%Performance on seen NEs is significantly and consistently higher than that of unseen NEs in different corpora, with the lower scores mostly attributable to lower recall.
%However, there are still significant differences at labelling seen NEs in different corpora.
%This means that NE diversity does not account for all of the difference of F1 between corpora of different genres.

To summarise, our findings are:
\begin{itemize}[noitemsep]
\item{F1 on unseen NEs is significantly lower than F1 on seen NEs for all three NERC approaches, which is mostly due to recall on unseen NEs being lower than that on seen NEs.}
\item{Performance on seen NEs is significantly and consistently higher than that of unseen NEs in different corpora, with the lower scores mostly attributable to lower recall.}
\item{However, there are still significant differences at labelling seen NEs in different corpora, which means that if NEs are seen or unseen does not account for all of the difference of F1 between corpora of different genres.}
\end{itemize}

\subsection{RQ4: Unseen Features, unseen NEs and NER performance}
% \item[RQ4] What is the relationship between Out-of-Vocabulary (OOV) features (unseen features), OOV entities (unseen NEs) and performance?

Having examined the impact of seen/unseen NEs on NERC performance in RQ3, and touched upon surface form drift in RQ2, we now turn our attention towards establishing the impact of seen features, i.e. features appearing in the test set that are observed also in the training set.
While feature sparsity can help to explain low F1, it is not a good predictor of performance across methods: sparse features can be good if mixed with high-frequency ones. For instance, Stanford NER often outperforms CRFSuite~(see Table~\ref{tab:F1}) despite having a lower proportion of seen features (i.e. those that occur both in test data and during training). Also, some approaches such as SENNA use a small number of features and base their features almost entirely on the NEs and not on their context.

\begin{figure}[!h]%[!htbp]
\centering
\includegraphics[width=1\textwidth]{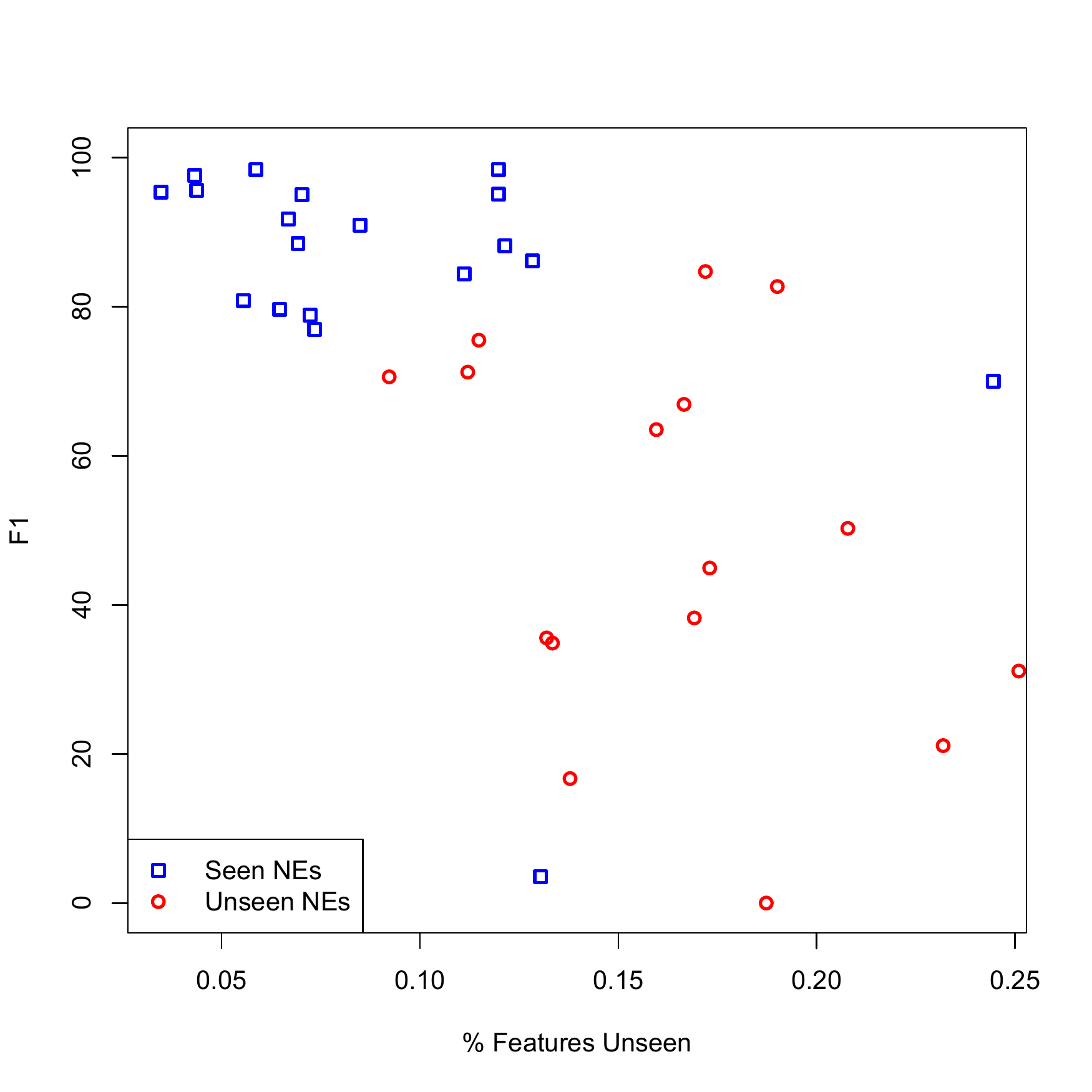}
\caption{Percentage of unseen features and F1 with Stanford NER for seen and unseen NEs in different corpora}\label{fig:PropFeatures}
\end{figure}

Subsequently, we want to measure F1 for unseens and seen NEs, as in Section~\ref{sec:models:sec:RQ3}, but also examine how the proportion of seen features impacts on the result.
We define seen features as those observed in the test data and also the training data.
In turn, unseen features are those observed in the test data but {\emph not} in the training data.
That is, they have not been previously encountered by the system at the time of labeling.
Unseen features are different from unseen words in that they are the difference in representation, not surface form.
For example, the entity ``Xoxarle" may be an unseen entity not found in training data
This entity could reasonably have ``shape:Xxxxxxx" and ``last-letter:e" as part of its feature representation.
If the training data contains entities ``Kenneth" and ``Simone", each of this will have generated these two features respectively.
Thus, these example features will not be unseen features in this case, despite coming from an unseen entity.
Conversely, continuing this example, if the training data contains no feature ``first-letter:X" -- which applies to the unseen entity in question -- then this {\emph will} be an unseen feature.

We therefore measure the proportion of unseen features per unseen and seen proportion of different corpora.
An analysis of this with Stanford NER is shown in Figure~\ref{fig:PropFeatures}. 
Each data point represents a corpus.
The blue squares are data points for seen NEs and the red circles are data points for unseen NEs. 
The figure shows a negative correlation between F1 and percentage of unseen features, i.e. the lower the percentage of unseen features, the higher the F1.
Seen and unseen performance and features separate into two groups, with only two outlier points.
The figure shows that novel, previously unseen NEs have more unseen features and that systems score a lower F1 on them.
This suggests that despite the presence of feature extractors for tackling unseen NEs, the features generated often do not overlap with those from seen NEs.
However, one would expect individual features to give different generalisation power for other sets of entities, and for systems use these features in different ways.
That is, machine learning approaches to the NER task do not seem to learn clear-cut decision boundaries based on a small set of features.
This is reflected in the softness of the correlation.

Finally, the proportion of seen features is higher for seen NEs. %The feature analysis is done with the Stanford NER features. SENNA is not very well suited for this kind of analysis because it uses mostly binary features based on embeddings, which are never ``unseen''.'
The two outlier points are ACE UN (low F1 for seen NEs despite low percentage of unseen features) and UMBC (high F1 for seen NEs despite high percentage of unseen features).
An error analysis shows that the ACE UN corpus suffers from the problem that the seen NEs are ambiguous, meaning even if they have been seen in the training corpus, a majority of the time they have been observed with a different NE label. For the UMBC corpus, the opposite is true: seen NEs are unambiguous.
This kind of metonymy is a known and challenging issue in NER, and the results on these corpora highlight the impact is still has on modern systems.
%We can conclude from this that ambiguity of seen NEs is also a factor in NER performance which should be taken into account.

For all approaches the proportion of observed features for seen NEs is bigger than the proportion of observed features for unseen NEs, as it should be. 
However, within the seen and unseen testing instances, there is no clear trend indicating whether having more observed features overall increases F1 performance. 
One trend that is observable is that the smaller the token/type ratio is (Table~\ref{tab:TokTypeRatio}), the bigger the variance between the smallest and biggest $n$ for each corpus, or, in other words, the smaller the token/type ratio is, the more diverse the features.

To summarise, our findings are:
\begin{itemize}[noitemsep]
\item{Seen NEs have more unseen features and systems score a lower F1 on them.}
\item{Outliers are due to low/high ambiguity of seen NEs.}
\item{The proportion of observed features for seen NEs is bigger than the proportion of observed features for unseen NEs}
\item{Within the seen and unseen testing instances, there is no clear trend indicating whether having more observed features overall increases F1 performance.}
\item{The smaller the token/type ratio is, the more diverse the features.}
\end{itemize}

\setlength{\tabcolsep}{0.2em}
\begin{table*}[t]%[!htbp]
\fontsize{9}{11}\selectfont
\centering
\begin{tabular}{c|c c c|c c c|c c c|c c c}
\toprule
& \multicolumn{3}{c}{\bf Memorisation} & \multicolumn{3}{c}{\bf CRFSuite} & \multicolumn{3}{c}{\bf Stanf} & \multicolumn{3}{c}{\bf SENNA} \\
& \bf P & \bf R & \bf F1 & \bf P & \bf R & \bf F1 & \bf P & \bf R & \bf F1 & \bf P & \bf R & \bf F1 \\
\hline
MUC 7 Dev & 38.24 & 20.89 & 27.02 & 54.27 & 50.09 & 52.09 & 57.01 & 55.42 & 56.21 & 50 & 59.97 & 54.53 \\
MUC 7 Test & 47.45 & 24.43 & 32.25 & 65.54 & 49.36 & 56.31 & 69.46 & 54.81 & 61.27 & 56.37 & 55.85 & 56.11 \\
\hline
CoNLL TestA & 53.14 & 22.36 & 31.48 & 67.12 & 38.57 & 48.99 & 69.22 & 48.27 & 56.88 & 68.62 & 58.68 & 63.26 \\
CoNLL TestB & 55.85 & 22.49 & 32.07 & 67.94 & 36.41 & 47.41 & 67.99 & 44.11 & 53.51 & 64.61 & 51.94 & 57.58 \\
\hline
ACE NW & 29.52 & 28.48 & 28.99 & 40.45 & 47.4 & 43.65 & 40.67 & 49.46 & 44.63 & 41.47 & 54 & 46.92 \\
ACE BN & 1.49 & 0.15 & 0.28 & 0 & 0 & 0 & 0 & 0 & 0 & 36.7 & 6.09 & 10.44 \\
ACE BC & 24.42 & 25.13 & 24.77 & 36.06 & 45.67 & 40.3 & 32.73 & 45.84 & 38.19 & 33.37 & 50.93 & 40.32 \\
ACE WL & 25.45 & 16.54 & 20.05 & 40.53 & 38.45 & 39.46 & 41.39 & 41.34 & 41.37 & 41.48 & 45.01 & 43.17 \\
ACE CTS & 68.31 & 25.58 & 37.23 & 26.28 & 16.94 & 20.6 & 35.93 & 22.47 & 27.65 & 24.69 & 23.05 & 23.84 \\
ACE UN & 8.07 & 27.69 & 12.5 & 9.76 & 40.08 & 15.7 & 10.48 & 42.56 & 16.82 & 9.95 & 49.59 & 16.57 \\
\hline
OntoNotes BN & 36.97 & 26.06 & 30.57 & 47.77 & 68.57 & 56.31 & 49.49 & 46.48 & 47.94 & 48.43 & 46.7 & 47.55 \\
OntoNotes BC & 33.68 & 24.21 & 28.17 & 72.24 & 64.74 & 68.29 & 72.69 & 66.47 & 69.44 & 69.49 & 70.88 & 70.18 \\
OntoNotes WB & 47.45 & 31.23 & 37.67 & 59.14 & 53.81 & 56.35 & 63.88 & 60.58 & 62.19 & 57.04 & 57.94 & 57.49 \\
OntoNotes TC & 54.15 & 28.4 & 37.26 & 60.88 & 48.26 & 53.84 & 65.09 & 60.1 & 62.5 & 57.79 & 62.02 & 59.83 \\
OntoNotes MZ & 40.38 & 20.1 & 26.84 & 47.75 & 64.05 & 54.71 & 51.31 & 41.05 & 45.61 & 43.23 & 39.05 & 41.03 \\
\hline
MSM 2013 Test & 14.87 & 5.8 & 8.34 & 41.29 & 23.32 & 29.81 & 49.2 & 32.19 & 38.92 & 16.81 & 37.85 & 23.28 \\
Ritter & 42.34 & 11.6 & 18.22 & 35.34 & 24.69 & 29.07 & 37.07 & 26.91 & 26.91 & 27.09 & 36.79 & 31.2 \\
UMBC & 52.27 & 12.17 & 19.74 & 44.71 & 20.21 & 27.84 & 59.09 & 27.51 & 37.55 & 31.39 & 22.75 & 26.38 \\
\hline\hline
{\bf Macro Average} & 35.48 & 19.65 & 23.87 & 43.00 & 38.45 & 38.99 & 45.93 & 40.29 & 41.45 & 40.98 & 43.64 & 40.51 \\
\bottomrule
\end{tabular}
\caption{\label{tab:F1OutOfDomain} Out of domain performance: F1 of NERC with different models}
\end{table*}

\subsection{RQ5: Out-Of-Domain NER Performance and Memorisation}\label{sec:RQ5}

This section explores baseline out-of-domain NERC performance without domain adaptation; what percentage of NEs are seen if there is a difference between the the training and the testing domains; and how the difference in performance on unseen and seen NEs compares to in-domain performance.

As demonstrated by the above experiments, and in line with related work, NERC performance varies across domains while also being influenced by the size of the available in-domain training data. Prior work on transfer learning and domain adaptation (e.g.~\cite{DaumeIII2007Domain}) has aimed at increasing performance in domains where only small amounts of training data are available. This is achieved by adding out-of domain data from domains where larger amounts of training data exist. For domain adaptation to be successful, however, the seed domain needs to be similar to the target domain, i.e. if there is no or very little overlap in terms of contexts of the training and testing instances, the model does not learn any additional helpful weights. As a confounding factor, Twitter and other social media generally consist of many (thousands-millions) of micro-domains, with each author~\cite{gella2014one} community~\cite{yang2015putting} and even conversation~\cite{pavalanathan2015audience} having its own style, which makes it hard to adapt to it as a single, monolithic genre; accordingly, adding out-of-domain NER data gives bad results in this situation~\cite{ritter2011named}. And even if recognised perfectly, entities that occur just once cause problems beyond NER, e.g. in co-reference~\cite{recasens2013life}.

In particular, \cite{sutton-mccallum:2005:HLTEMNLP} has reported improving F1 by around 6\% through adaptation from the CoNLL to the ACE dataset. However, transfer learning becomes more difficult if the target domain is very noisy or, as mentioned already, too different from the seed domain. For example,~\cite{locke2009named} unsuccessfully tried to adapt the CoNLL 2003 corpus to a Twitter corpus spanning several topics. They found that hand-annotating a Twitter corpus consisting of 24,000 tokens performs better on new Twitter data than their transfer learning efforts with the CoNLL 2003 corpus.

The seed domain for the experiments here is newswire, where we use the classifier trained on the biggest NW corpus investigated in this study, i.e. OntoNotes NW. That classifier is then applied to all other corpora. The rationale is to test how suitable such a big corpus would be for improving Twitter NER, for which only small training corpora are available.

%The research questions investigated in this section are: how does out-of-domain performance compare with in-domain performance on the same corpora and what conclusions can be drawn from that? Which domains or corpora have particularly high or low out-of-domain performance?

Results for out-of-domain performance are reported in Table~\ref{tab:F1OutOfDomain}. The highest F1 performance is on the OntoNotes BC corpus, with similar results to the in-domain task. This is unsurprising as it belongs to a similar domain as the training corpus (broadcast conversation) the data was collected in the same time period, and it was annotated using the same guidelines. In contrast, out-of-domain results are much lower than in-domain results for the CoNLL corpora, even though they belong to the same genre as OntoNotes NW. Memorisation recall performance on CoNLL TestA and TestB with OntoNotes NW test suggest that this is partly due to the relatively low overlap in NEs between the two datasets. This could be attributed to the CoNLL corpus having been collected in a different time period to the OntoNotes corpus, when other entities were popular in the news; an example of drift~\cite{masud2010addressing}. Conversely, Stanford NER does better on these corpora than it does on other news data, e.g. ACE NW. This indicates that Stanford NER is capable of some degree of generalisation and can detect novel entity surface forms; however, recall is still lower than precision here, achieving roughly the same scores across these three (from 44.11 to 44.96), showing difficulty in picking up novel entities in novel settings.

In addition, there are differences in annotation guidelines between the two datasets. If the CoNLL annotation guidelines were more inclusive than the Ontonotes ones, then even a memorisation evaluation over the same dataset would yield this result. This is, in fact, the case: OntoNotes divides entities into more classes, not all of which can be readily mapped to PER/LOC/ORG. For example, OntoNotes includes PRODUCT, EVENT, and WORK OF ART classes, which are not represented in the CoNLL data. It also includes the NORP class, which blends nationalities, religious and political groups. This has some overlap with ORG, but also includes terms such as ``muslims" and ``Danes", which are too broad for the ACE-related definition of ORGANIZATION. Full details can be found in the OntoNotes 5.0 release notes\footnote{https://catalog.ldc.upenn.edu/docs/LDC2013T19/OntoNotes-Release-5.0.pdf , Section 2.6} and the (brief) CoNLL 2003 annotation categories.\footnote{http://www.cnts.ua.ac.be/conll2003/ner/annotation.txt} Notice how the CoNLL guidelines are much more terse, being generally non-prose, but also manage to cram in fairly comprehensive lists of sub-kinds of entities in each case. This is likely to make the CoNLL classes include a diverse range of entities, with the many suggestions acting as generative material for the annotator, and therefore providing a broader range of annotations from which to generalise from -- i.e., slightly easier to tag.

%\todo{Is there anything useful to say about P and R or only about F1?}
%\ld{the tables would be a lot friendlier if we reported just F1 - but maybe that's best left for the reviewers to comment on}

The lowest F1 of 0 is ``achieved" on ACE BN. An examination of that corpus reveals the NEs contained in that corpus are all lower case, whereas those in OntoNotes NW have initial capital letters.

\setlength{\tabcolsep}{0.2em}
\begin{table*}%[!htbp] % works as sidewaystable too, but for submission, we'll need to decrease the font size
\fontsize{8}{10}\selectfont
\centering
\begin{tabular}{c c|c c c|c c c|c c c|c c c}
\toprule
& &  \multicolumn{3}{c}{\bf CRFSuite} & \multicolumn{3}{c}{\bf Stanf} & \multicolumn{3}{c}{\bf SENNA} & \multicolumn{3}{c}{\bf All} \\
&  & \bf P & \bf R & \bf F1 & \bf P & \bf R & \bf F1 & \bf P & \bf R & \bf F1 & \bf P & \bf \bf R & \bf F1 \\
\hline\hline
MUC 7 Dev & Seen & 81.25 & 55.15 & 65.70 & 82.1 & 56.97 & 67.26 & 86.21 & 68.18 & 76.14 & 83.19 & 60.10 & 69.70 \\
& Unseen & 63.40 & 50.83 & 56.42 & 72.22 & 59.09 & 65.00 & 64.79 & 57.02 & 60.66 & 66.80 & 55.65 & 60.69 \\
\hline
MUC 7 Test & Seen & 81.25 & 54.93 & 65.55 & 79.15 & 52.72 & 63.29 & 82.43 & 61.37 & 70.36 & 80.94 & 56.34 & 66.40 \\
& Unseen & 65.37 & 50.55 & 57.01 & 77.78 & 65.03 & 70.83 & 62.71 & 60.66 & 61.67 & 68.62 & 58.75 & 63.17 \\
\hline\hline
CoNLL TestA & Seen & 72.49 & 35.79 & 47.92 & 72.33 & 44.78 & 55.31 & 78.77 & 58.77 & 67.31 & 74.53 & 46.45 & 56.85 \\
& Unseen & 79.32 & 49.63 & 61.06 & 82.61 & 60.9 & 70.12 & 76.96 & 65.2 & 70.59 & 79.63 & 58.58 & 67.26 \\
\hline
CoNLL TestB & Seen & 74.72 & 35.97 & 48.56 & 73.3 & 43.08 & 54.27 & 74.32 & 52.77 & 61.71 & 74.11 & 43.94 & 54.85 \\
& Unseen & 75.38 & 42.39 & 54.27 & 76.18 & 53.04 & 62.54 & 68.76 & 56.03 & 61.75 & 73.44 & 50.49 & 59.52 \\
\hline\hline
ACE NW & Seen & 76.21 & 50.45 & 60.71 & 79.32 & 54.03 & 64.28 & 86.07 & 61.72 & 71.89 & 80.53 & 55.40 & 65.63 \\
& Unseen & 46.70 & 46.05 & 46.37 & 45.18 & 45.81 & 45.50 & 43.38 & 47.21 & 45.21 & 45.09 & 46.36 & 45.69 \\
\hline
ACE BN & Seen & 0.00 & 0.00 & 0.00 & 0 & 0 & 0.00 & 96.67 & 8.19 & 15.1 & 32.22 & 2.73 & 5.03 \\
& Unseen & 0.00 & 0.00 & 0.00 & 0 & 0 & 0.00 & 36.11 & 4.29 & 7.67 & 12.04 & 1.43 & 2.56 \\
\hline
ACE BC & Seen & 82.11 & 52.65 & 64.16 & 82.43 & 53.82 & 65.12 & 88.98 & 61.76 & 72.92 & 84.51 & 56.08 & 67.40 \\
& Unseen & 39.92 & 38.15 & 39.01 & 42.44 & 40.56 & 41.48 & 41.25 & 42.57 & 41.9 & 41.20 & 40.43 & 40.80 \\
\hline
ACE WL & Seen & 66.00 & 41.04 & 50.61 & 68.82 & 45.02 & 54.44 & 48.2 & 44.72 & 64.30 & 61.01 & 43.59 & 56.45 \\
& Unseen & 45.79 & 37.78 & 41.40 & 47.39 & 40.28 & 43.54 & 79.49 & 53.98 & 46.40 & 57.56 & 44.01 & 43.78 \\
\hline
ACE CTS & Seen & 91.75 & 46.55 & 61.76 & 82.74 & 41.59 & 55.35 & 87.13 & 61.55 & 72.14 & 87.21 & 49.90 & 63.08 \\
& Unseen & 54.69 & 49.30 & 51.85 & 58.46 & 53.52 & 55.88 & 54.41 & 52.11 & 53.24 & 55.85 & 51.64 & 53.66 \\
\hline
ACE UN & Seen & 74.51 & 44.71 & 55.88 & 75.93 & 48.24 & 58.99 & 90.99 & 59.41 & 71.89 & 80.48 & 50.79 & 62.25 \\
& Unseen & 37.50 & 29.17 & 32.81 & 43.48 & 27.78 & 33.90 & 33.93 & 26.39 & 29.69 & 38.30 & 27.78 & 32.13 \\
\hline\hline
OntoNotes BN & Seen & 63.92 & 53.09 & 58.00 & 66.06 & 56.71 & 61.03 & 66.8 & 58.73 & 62.50 & 65.59 & 56.18 & 60.51 \\
& Unseen & 35.42 & 32.42 & 33.85 & 36.39 & 34.33 & 35.33 & 34.13 & 32.31 & 33.20 & 35.31 & 33.02 & 34.13 \\
\hline
OntoNotes BC & Seen & 84.83 & 66.05 & 74.27 & 86.41 & 70.08 & 77.39 & 87.58 & 75.85 & 81.29 & 86.27 & 70.66 & 77.65 \\
& Unseen & 76.74 & 65.54 & 70.70 & 82 & 72.39 & 76.90 & 71.95 & 68.02 & 69.93 & 76.90 & 68.65 & 72.51 \\
\hline
OntoNotes WB & Seen & 75.44 & 58.07 & 65.62 & 79.64 & 65.23 & 71.71 & 79.75 & 64.08 & 71.06 & 78.28 & 62.46 & 69.46 \\
& Unseen & 61.37 & 47.93 & 53.82 & 67.89 & 54.47 & 60.44 & 55.22 & 49.4 & 52.15 & 61.49 & 50.60 & 55.47 \\
\hline
OntoNotes TC & Seen & 71.33 & 48.89 & 58.02 & 76.19 & 62.9 & 68.91 & 84.57 & 70.02 & 76.61 & 77.36 & 60.60 & 67.85 \\
& Unseen & 67.72 & 51.50 & 58.50 & 75.57 & 59.28 & 66.44 & 58.7 & 48.5 & 53.11 & 67.33 & 53.09 & 59.35 \\
\hline
OntoNotes MZ & Seen & 64.84 & 46.61 & 54.24 & 64.34 & 48.23 & 55.13 & 61.7 & 46.53 & 53.05 & 63.63 & 47.12 & 54.14 \\
& Unseen & 41.40 & 28.93 & 34.06 & 49.7 & 32.63 & 39.40 & 38.92 & 30.43 & 34.16 & 43.34 & 30.66 & 35.87 \\
\hline\hline
MSM 2013 Test & Seen & 58.90 & 19.24 & 29.01 & 56.25 & 22.15 & 31.78 & 57.08 & 30.65 & 39.88 & 57.41 & 24.01 & 33.56 \\
& Unseen & 70.30 & 35.33 & 47.03 & 73.5 & 45.89 & 56.50 & 59.12 & 48.02 & 53.00 & 67.64 & 43.08 & 52.18 \\
\hline
Ritter & Seen & 62.75 & 25.20 & 35.96 & 58.77 & 26.38 & 36.41 & 79.69 & 40.16 & 53.40 & 67.07 & 30.58 & 41.92 \\
& Unseen & 58.90 & 28.48 & 38.39 & 56.47 & 31.79 & 40.68 & 62.5 & 39.74 & 48.58 & 59.29 & 33.34 & 42.55 \\
\hline
UMBC & Seen & 60.53 & 20.18 & 30.26 & 75 & 26.09 & 38.71 & 72.34 & 29.57 & 41.98 & 69.29 & 25.28 & 36.98 \\
& Unseen & 60.61 & 27.03 & 37.38 & 73.53 & 33.78 & 46.30 & 38.3 & 24.32 & 29.75 & 57.48 & 28.38 & 37.81 \\
\hline\hline
{\bf Macro Average} & Seen & 69.05 & 41.92 & 51.46 & 69.93 & 45.45 & 54.41 & 78.29 & 53.00 & 62.42 & 72.42 & 46.79 & 56.10 \\
& Unseen & 54.47 & 39.50 & 45.22 & 58.93 & 45.03 & 50.60 & 54.48 & 44.79 & 47.37 & 55.96 & 43.11 & 47.73 \\
\bottomrule
\end{tabular}
\caption{\label{tab:F1OutOfDomainSing} Out-of-domain performance for unseen vs. seen NEs: F1 of NERC with different models}
\end{table*}

Results on unseen NEs for the out-of-domain setting are in Table~\ref{tab:F1OutOfDomainSing}. The last section's observation of NERC performance being lower for unseen NEs also generally holds true in this out-of-domain setting. The macro average over F1 for the in-domain setting is 76.74\% for seen NEs vs. 53.76 for unseen NEs, whereas for the out-of-domain setting the F1 is 56.10\% for seen NEs and 47.73\% for unseen NEs.

Corpora with a particularly big F1 difference between seen and unseen NEs ($<=$ 20\% averaged over all NERC methods) are ACE NW, ACE BC, ACE UN, OntoNotes BN and OntoNotes MZ. For some corpora (CoNLL Test A and B, MSM and Ritter), out-of-domain F1 (macro average over all methods) of unseen NEs is better than for seen NEs. We suspect that this is due to the out-of-domain evaluation setting encouraging better generalisation, as well as the regularity in entity context observed in the fairly limited CoNLL news data -- for example, this corpus contains a large proportion of cricket score reports and many cricketer names, occurring in linguistically similar contexts. Others have also noted that the CoNLL datasets are low-diversity compared to OntoNotes, in the context of named entity recognition~\cite{chiu2016named}. In each of the exceptions except MSM, the difference is relatively small. We note that the MSM test corpus is one of the smallest datasets used in the evaluation, also based on a noisier genre than most others, and so regard this discrepancy as an outlier.

Corpora for which out-of-domain F1 is better than in-domain F1 for at least one of the NERC methods are: MUC7 Test, ACE WL, ACE UN, OntoNotes WB, OntoNotes TC and UMBC. Most of those corpora are small, with combined training and testing bearing fewer than 1,000 NEs (MUC7 Test, ACE UN, UMBC). In such cases, it appears beneficial to have a larger amount of training data, even if it is from a different domain and/or time period. The remaining 3 corpora contain weblogs (ACE WL, ACE WB) and online Usenet discussions (ACE UN). Those three are diverse corpora, as can be observed by the relatively low NEs/Unique NEs ratios~(Table~\ref{tab:TokTypeRatioNE}). However, NE/Unique NEs ratios are not an absolute predictor for better out-of-domain than in-domain performance: there are corpora with lower NEs/Unique NEs ratios than ACE WB which have better in-domain than out-of-domain performance. As for the other Twitter corpora, MSM 2013 and Ritter, performance is very low, especially for the memorisation system. This reflects that, as well as surface form variation, the context or other information represented by features shifts significantly more in Twitter than across different samples of newswire, and that the generalisations that can be drawn from newswire by modern NER systems are not sufficient to give any useful performance in this natural, unconstrained kind of text.

In fact, it is interesting to see that the memorisation baseline is so effective with many genres, including broadcast news, weblog and newswire.
This indicates that there is low variation in the topics discussed by these sources -- only a few named entities are mentioned by each.
When named entities are seen as micro-topics, each indicating a grounded and small topic of interest, this reflects the nature of news having low topic variation, focusing on a few specific issues -- e.g., location referred to tend to be big; persons tend to be politically or financially significant; and organisations rich or governmental~\cite{Bontcheva2014}.
In contrast, social media users also discuss local locations like restaurants, organisations such as music band and sports clubs, and are content to discuss people that are not necessarily mentioned in Wikipedia.
The low overlap and memorisation scores on tweets, when taking entity lexica based on newswire, are therefore symptomatic of the lack of variation in newswire text, which has a limited authorship demographic~\cite{eisenstein2013bad} and often has to comply to editorial guidelines.

The other genre that was particularly difficult for the systems was ACE Usenet.
This is a form of user-generated content, not intended for publication but rather discussion among communities.
In this sense, it is social media, and so it is not surprising that system performance on ACE UN resembles performance on social media more than other genres.

Crucially, the computationally-cheap memorisation method actually acts as a reasonable predictor of the performance of other methods.
This suggests that high entity diversity predicts difficulty for current NER systems.
As we know that social media tends to have high entity diversity -- certainly higher than other genres examined -- this offers an explanation for why NER systems perform so poorly when taken outside the relatively conservative newswire domain.
Indeed, if memorisation offers a consistent prediction of performance, then it is reasonable to say that memorisation and memorisation-like behaviour accounts for a large proportion of NER system performance.

To conclude regarding memorisation and out-of-domain performance, there are multiple issues to consider: is the corpus a sub-corpus of the same corpus as the training corpus, does it belong to the same genre, is it collected in the same time period, and was it created with similar annotation guidelines.
Yet it is very difficult to explain high/low out-of-domain performance compared to in-domain performance with those factors.

A consistent trend is that, if out-of-domain memorisation is better in-domain memorisation, out-of-domain NERC performance with supervised learning is better than in-domain NERC performance with supervised learning too. This reinforces discussions in previous sections: an overlap in NEs is a good predictor for NERC performance.
This is useful when a suitable training corpus has to be identified for a new domain. It can be time-consuming to engineer features or study and compare machine learning methods for different domains, while memorisation performance can be checked quickly.

Indeed, memorisation consistently predicts NER performance.
The prediction applies both within and across domains.
This has implications for the focus of future work in NER: the ability to generalise well enough to recognise unseen entities is a significant and still-open problem.

%\ia{Write more discussion for seen/unseen NEs for out of domain setting. We have show results for this but don't discuss them much.}

To summarise, our findings are:
\begin{itemize}[noitemsep]
\item{What time period an out of domain corpus is collected in plays an important role in NER performance.}
\item{The context or other information represented by features shifts significantly more in Twitter than across different samples of newswire.}
\item{The generalisations that can be drawn from newswire by modern NER systems are not sufficient to give any useful performance in this varied kind of text.}
\item{Memorisation consistently predicts NER performance, both inside and outside genres or domains.}
\end{itemize}

\section{Conclusion}
\label{sec:conclusion}

This paper investigated the ability of modern NER systems to generalise effectively over a variety of genres.
Firstly, by analysing different corpora, we demonstrated that datasets differ widely in many regards: in terms of size; balance of entity classes; proportion of NEs; and how often NEs and tokens are repeated. 
The most balanced corpus in terms of NE classes is the CoNLL corpus, which, incidentally, is also the most widely used NERC corpus, both for method tuning of off-the-shelf NERC systems (e.g. Stanford NER, SENNA), as well as for comparative evaluation. 
Corpora, traditionally viewed as noisy, i.e. the Twitter and Web corpora, were found to have a low repetition of NEs and tokens. More surprisingly, however, so does the CoNLL corpus, which indicates that it is well balanced in terms of stories. 
Newswire corpora have a large proportion of NEs as percentage of all tokens, which indicates high information density. 
Web, Twitter and telephone conversation corpora, on the other hand, have low information density.

Our second set of findings relates to the NERC approaches studied. Overall, SENNA achieves consistently the highest performance across most corpora, and thus has the best approach to generalising from training to testing data. 
This can mostly be attributed to SENNA's use of word embeddings, trained with deep convolutional neural nets. 
The default parameters of SENNA achieve a balanced precision and recall, while for Stanford NER and CRFSuite, precision is almost twice as high as recall. 
 
Our experiments also confirmed the correlation between NERC performance and training corpus size, although size alone is not an absolute predictor. 
In particular, the biggest NE-annotated corpus amongst those studied is OntoNotes NW -- almost twice the size of CoNLL in terms of number of NEs. 
Nevertheless, the average F1 for CoNLL is the highest of all corpora and, in particular, SENNA has 11 points higher F1 on CoNLL than on OntoNotes NW.

Studying NERC on size-normalised corpora, it becomes clear that there is also a big difference in performance on corpora from the same genre. 
When normalising training data by size, diverse corpora, such as Web and social media, still yield lower F1 than newswire corpora. 
This indicates that annotating more training examples for diverse genres would likely lead to a dramatic increase in F1.

What is found to be a good predictor of F1 is a memorisation baseline, which picks the most frequent NE label for each token sequence in the test corpus as observed in the training corpus. 
This supported our hypothesis that entity diversity plays an important role, being negatively correlated with F1.
Studying proportions of unseen entity surface forms, experiments showed corpora with a large proportion of unseen NEs tend to yield lower F1, due to much lower performance on unseen than seen NEs (about 17 points lower averaged over all NERC methods and corpora). 
This finally explains why the performance is highest for the benchmark CoNLL newswire corpus -- it contains the lowest proportion of unseen NEs.
It also explains the difference in performance between NERC on other corpora. 
Out of all the possible indicators for high NER F1 studied, this is found to be the most reliable one.
This directly supports our hypothesis that generalising for unseen named entities is both difficult and important.

Also studied is the proportion of unseen features per unseen and seen NE portions of different corpora. 
However, this is found to not be very helpful. 
The proportion of seen features is higher for seen NEs, as it should be. 
However, within the seen and unseen NE splits, there is no clear trend indicating if having more seen features helps.

We also showed that hand-annotating more training examples is a straight-forward and reliable way of improving NERC performance. 
However, this is costly, which is why it can be useful to study if using different, larger corpora for training might be helpful. 
Indeed, substituting in-domain training corpora with other training corpora for the same genre created at the same time improves performance, and studying how such corpora can be combined with transfer learning or domain adaptation strategies might improve performance even further. 
However, for most corpora, there is a significant drop in performance for out-of-domain training. 
What is again found to be reliable is to check the memorisation baseline: if results for the out-of-domain memorisation baseline are higher than for in-domain memorisation, than using the out-of-domain corpus for training is likely to be helpful.

Across a broad range of corpora and genres, characterised in different ways, we have examined how named entities are embedded and presented.
While there is great variation in the range and class of entities found, it is consistent that the more varied texts are harder to do named entity recognition in.
This connection with variation occurs to such an extent that, in fact, performance when memorising lexical forms stably predicts system accuracy.
The result of this is that systems are not sufficiently effective at generalising beyond the entity surface forms and contexts found in training data.
To close this gap and advance NER systems, and cope with the modern reality of streamed NER, as opposed to the prior generation of batch-learning based systems with static evaluation sets being used as research benchmarks, future work needs to address named entity generalisation and out-of-vocabulary lexical forms.

\section*{Acknowledgement}
This work was partially supported by the UK EPSRC Grant No. EP/K017896/1 uComp\footnote{\url{http://www.ucomp.eu/}} and by the European Union under Grant Agreements No. 611233 PHEME.\footnote{\url{http://www.pheme.eu/}}
The authors wish to thank the CS\&L reviewers for their helpful and constructive feedback.

\bibliographystyle{abbrvnat}
\bibliography{ner-generalisation}

\end{document}